\documentclass[10pt,journal,compsoc]{IEEEtran}
\usepackage{amsmath,amsfonts}
\usepackage{algorithmic}
\usepackage{array}
\usepackage{algorithm}
\usepackage[pagebackref,breaklinks,colorlinks]{hyperref}
\usepackage{hyperref}
\hypersetup{
    colorlinks=true,
    linkcolor=blue,
    filecolor=blue,      
    urlcolor=blue,
    citecolor=cyan,
}

\usepackage[table,xcdraw,dvipsnames]{xcolor}
\usepackage[caption=false,font=normalsize,labelfont=sf,textfont=sf]{subfig}
\usepackage{textcomp}
\usepackage[font=small,labelfont=bf]{caption}
\usepackage{stfloats}
\usepackage{multirow}
\usepackage{bbding}
\usepackage{url}
\usepackage{cite}
\usepackage{verbatim}
\usepackage{graphicx}
\usepackage{colortbl}
\usepackage{enumitem}
\usepackage{booktabs}
\usepackage{balance}
\usepackage{ragged2e}
\definecolor{perception}{RGB}{248,210,119}
\definecolor{planning}{RGB}{192,214,235}
\definecolor{action}{RGB}{159,206,99}
\definecolor{reflection}{RGB}{180,143,218}
\definecolor{correct}{RGB}{197,224,180}
\definecolor{unnec}{RGB}{243,206,133}
\definecolor{wrong}{RGB}{255,142,133}
\definecolor{green}{RGB}{84,130,53}
\usepackage{listings}
\usepackage{nicefrac}      
\usepackage{microtype}      
\usepackage{multirow}
\usepackage{array}
\usepackage{colortbl}
\usepackage{enumitem}
\usepackage{wrapfig}
\usepackage{amssymb}
\usepackage{xcolor}        
\usepackage{tabularx}
\usepackage{tcolorbox}
\tcbuselibrary{breakable}
\usepackage{arydshln}
\usepackage{makecell}

\begin{document}

\title{Optimus-3: Dual-Router Aligned Mixture-of-Experts Agent with Dual-Granularity Reasoning-Aware Policy Optimization}

\author{Zaijing Li, Rui Shao,~\IEEEmembership{Member,~IEEE}, Yuquan Xie, Gongwei Chen, Weili Guan, \\
Dongmei Jiang, Yaowei Wang,~\IEEEmembership{Member,~IEEE}, and Liqiang Nie,~\IEEEmembership{Senior Member,~IEEE}
\IEEEcompsocitemizethanks{
\IEEEcompsocthanksitem Zaijing Li and Yaowei Wang are with the School of Computer Science and Technology, Harbin Institute of Technology (Shenzhen Campus), Shenzhen 518055, China and Pengcheng Laboratory, Shenzhen 518000, China\\
E-mail: lzj14011@gmail.com, wangyw@pcl.ac.cn
\IEEEcompsocthanksitem Rui Shao, Yuquan Xie, and Liqiang Nie are with School of Computer Science and Technology, Harbin Institute of Technology (Shenzhen Campus), Shenzhen 518055, China \\
E-mail: rshaojimmy@gmail.com, nieliqiang@gmail.com
\IEEEcompsocthanksitem Weili Guan and Gongwei Chen are with the School of Information Science and Technology, Harbin Institute of Technology (Shenzhen Campus), Shenzhen 518055, China and Pengcheng Laboratory, Shenzhen, China and Shenzhen Loop Area Institute, Shenzhen, China\\
E-mail: honeyguan@gmail.com, cgwfeel@163.com
\IEEEcompsocthanksitem Dongmei Jiang is with Pencheng Laboratory, Shenzhen 518000, China \\
E-mail: jiangdm@pcl.ac.cn
\IEEEcompsocthanksitem The corresponding authors are Rui Shao and Liqiang Nie.}\\
\texttt{{\url{https://cybertronagent.github.io/Optimus-3.github.io/}}}
}

\IEEEtitleabstractindextext{
\begin{abstract}
\justifying
Developing generalist agents capable of solving open-ended tasks in visually rich, dynamic environments remains a core pursuit of embodied AI. While Minecraft has emerged as a compelling benchmark, existing agents often suffer from fragmented cognitive abilities, lacking the synergy between reflexive execution (\textbf{System 1}) and deliberative reasoning (\textbf{System 2}). In this paper, we introduce \textbf{Optimus-3}, a generalist agent that organically integrates these dual capabilities within a unified framework. To achieve this, we address three fundamental challenges. First, to overcome the scarcity of reasoning data, we propose a \textbf{Knowledge-Enhanced Automated Data Generation Pipeline}. It synthesizes high-quality System 2 reasoning traces from raw System 1 interaction trajectories, effectively mitigating hallucinations via injection of domain knowledge. We release the resulting dataset, \textbf{OptimusM$^{4}$}, to the community. Second, to reconcile the dichotomous computational requirements of the dual systems, we design a \textbf{Dual-Router Aligned MoE Architecture}. It employs a Task Router to prevent task interference via parameter decoupling, and a Layer Router to dynamically modulate reasoning depth, creating a computational ``Fast Path'' for System 1 and a ``Deep Path'' for System 2. Third, to activate the reasoning capabilities of System 2, we propose \textbf{Dual-Granularity Reasoning-Aware Policy Optimization (DGRPO)} algorithm. It enforces Process-Outcome Co-Supervision via dual-granularity dense rewards, ensuring consistency between the thought process and the answer. Extensive evaluations demonstrate that Optimus-3 surpasses existing state-of-the-art methods on both System~2 (21$\%$ on Planning, 66\% on Captioning, 76\% on Embodied QA, 3.4$\times$ on Grounding, and 18\% on Reflection) and System~1 (3\% on Long-Horizon Action) tasks, with a notable 60\% success rate on open-ended tasks.

\end{abstract}

}

\maketitle

\IEEEdisplaynontitleabstractindextext

\section{Introduction}
Building generalist agents that can solve open-ended tasks in visually rich, open-world environments is a long-standing vision of embodied AI \cite{hafner2023mastering,tan2024towards,raad2024scaling,yang2024video}. Among such environments, Minecraft \cite{guss2019minerl,fan2022minedojo} has emerged as a compelling benchmark for studying open-world agents, due to its diverse scenes and objects, vast action space, and the open-ended nature of tasks. Recent advances in Minecraft \cite{wang2023voyager,qin2023mp5,wang2023jarvis,wang2023describe,li2024optimus,li2025optimus} have empowered agents with Multimodal Large Language Models (MLLMs) \cite{bai2025qwen2,chen2024lion,liu2024visual}, enabling impressive performance on programmatic, pre-defined objectives. However, these agents often exhibit brittle behavior when confronted with dynamic and open-ended instructions, where success necessitates both immediate responsiveness to dynamic changes and deliberate analysis of visual cues.

\begin{figure*}[htbp]
    \centering
    \includegraphics[width=1\textwidth]{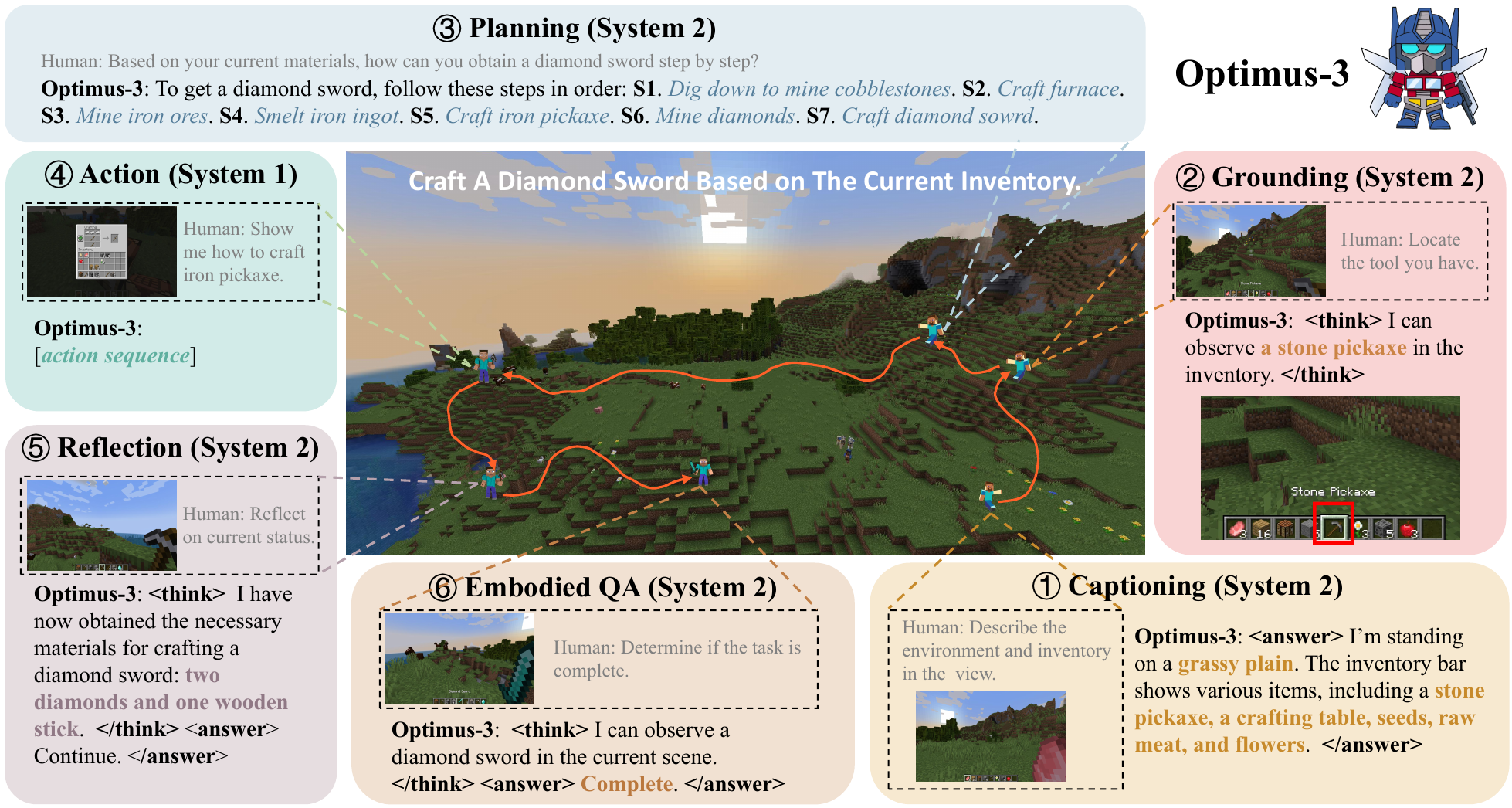}
    \caption{Given the task \textit{Craft a diamond sword based on the current inventory}, Optimus-3 employs \textbf{Captioning} to perceive and interpret the inventory information, \textbf{Grounding} to select appropriate tools, \textbf{Planning} to generate sub-goals based on available materials, \textbf{Action} to execute these sub-goals sequentially, \textbf{Reflection} to assess the current task state, and \textbf{Embodied QA} to verify whether the task has been successfully completed.}
    \label{fig:fig1}
\end{figure*}

To achieve this, we argue that an agent must emulate the \textbf{Dual-Process Theory} of human cognition \cite{kahneman2011thinking}: synergizing System 1 (fast, reflexive, intuitive) and System 2 (slow, deliberative, analytic) processes. As illustrated in Figure~\ref{fig:fig1}, consider the instruction \emph{``Craft a diamond sword based on the current inventory''}. This open-ended task demands a seamless interplay between two cognitive modes. \textbf{System 2} takes charge of the deliberative reasoning loop, encompassing Planning, Reflection, and notably, Active Perception (Grounding, Embodied QA). We classify these visual tasks as System 2 processes because they transcend passive detection: they demand a deliberative analysis of visual evidence to verify the factual existence of queried entities before decision-making. \textbf{System 1} governs the high-frequency visuomotor loop, executing low-latency control commands (Action) to interact with the environment. Existing Minecraft agents, however, typically possess only a fragmented subset of these abilities, lacking the organic synergy required for robust performance. This raises a fundamental research question: \textbf{Can we develop a generalist agent that organically integrates System 1 action loops with System 2 reasoning capabilities within a unified framework?} Toward this goal, we identify three fundamental challenges centered around the acquisition, modeling, and activation of these dual capabilities.

\textbf{Challenge \uppercase\expandafter{\romannumeral 1}: Scarcity of Domain-Specific System 2 Reasoning Traces.}  While raw gameplay data (action trajectories) is abundant \cite{vpt,lifshitz2024steve}, there is a critical scarcity of data capturing the reasoning processes, such as hierarchical planning, visual grounding, and reflection. Manually annotating these complex cognitive traces is prohibitively expensive. A naive strategy is to leverage general-domain MLLMs \cite{liu2024visual,chen2024lion,bai2025qwen2} for automated annotation. However, these models, while possessing strong general reasoning, lack the ``knowledge'' of Minecraft (e.g., crafting recipes, physics rules). Consequently, they frequently generate unfeasible plans or hallucinated responses \cite{bai2025qwen2,lu2024deepseek}. Bridging the gap between general-purpose reasoning and domain-specific knowledge to construct high-quality System 2 reasoning traces remains a significant hurdle.

\textbf{Challenge \uppercase\expandafter{\romannumeral 2}: Computational Conflicts in Coupling System 1 and System 2.}  System 1 (Action) demands high-frequency inference with low latency, relying on shallow, reflexive processing of local context. In contrast, System 2 (Planning, Reflection) operates at a lower frequency but requires deep, computation-intensive reasoning over long horizons. Standard dense architectures or conventional Mixture-of-Experts (MoE) \cite{moesurvey,huang2025modes}  enforce a uniform computational depth, resulting in a dilemma: they are either too computationally heavy for real-time control or too shallow for complex reasoning. Furthermore, training these heterogeneous tasks within a shared parameter space often leads to task interference \cite{shen2025mome,dai2401deepseekmoe}. Designing an architecture that adaptively allocates computational resources, provides a \emph{fast path} for System 1 and a \emph{deep path} for System 2, and simultaneously avoids task interference is highly challenging.

\begin{figure*}[htbp]
    \centering
    \includegraphics[width=1.0\textwidth]{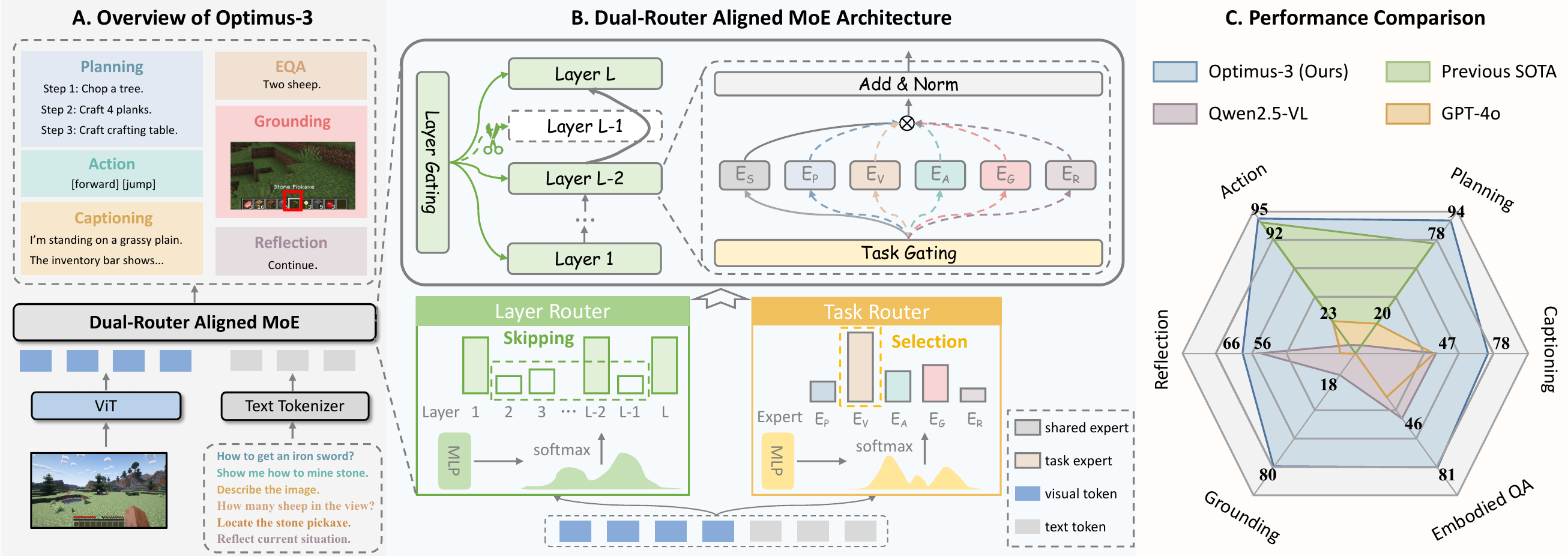}
    \caption{\textbf{(A)}: Overview of Optimus-3. Given observations and instructions, Optimus-3 couples System-1 fast reaction (Action) and System-2 deliberate reasoning (Embodied QA, Planning, Grounding, Reflection) within the Dual-Router Aligned MoE architecture. \textbf{(B)}: The details of Dual-Router Aligned MoE architecture. Horizontally, Task Router assigns each input to its corresponding task expert together with a shared knowledge expert. Vertically, Layer Router accelerates latency-sensitive action inference by selectively skipping intermediate layers. Both routing decisions are made once before the forward pass. \textbf{(C)}: Performance comparison of Optimus-3 against current task-specific SOTA agents, GPT-4o \cite{gpt4}, and Qwen2.5-VL \cite{bai2025qwen2}.}
    \label{fig:fig2}
\end{figure*}
\textbf{Challenge \uppercase\expandafter{\romannumeral 3}: Optimizing System 2 Reasoning under Sparse Outcome Feedback.} Given Minecraft's infinite diversity in scenes and dynamics, Supervised Fine-Tuning (SFT) is insufficient for agent training. While Reinforcement Learning (RL) \cite{schulman2017ppo,rafailov2023dpo} offers a pathway to activate such reasoning capabilities through exploration \cite{guo2025deepseek,shao2024deepseekmath}, adapting it to Minecraft presents distinct challenges. (i) Visual-Logic Misalignment: Unlike text-only tasks, Minecraft requires reasoning over high-dimensional visual inputs. Existing methods \cite{schulman2017ppo,rafailov2023dpo,shao2024deepseekmath} often struggle to ground reasoning in visual evidence, leading to perceptual hallucinations. (ii) Lack of Process Supervision: Standard RL typically relies on sparse outcome rewards, treating the reasoning process as a black box. This coarse supervision fails to penalize flawed intermediate logic that accidentally yields correct answers, hindering the acquisition of strictly valid and grounded reasoning chains.

In this paper, we propose \textbf{Optimus-3}, a generalist agent in open-world of Minecraft, which is endowed with comprehensive capabilities in perception, planning, action, and reflection (depicted in Figure \ref{fig:fig2}). To address the aforementioned challenges, we propose targeted improvements across three key dimensions: data generation, model architecture, and training methodology.

\textbf{Knowledge-Enhanced Data Generation Pipeline for System 2 Tasks.} To address the scarcity of domain-specific reasoning data (Challenge \uppercase\expandafter{\romannumeral 1}), we introduce a knowledge-enhanced pipeline that synthesizes high-quality  reasoning traces from raw action trajectories. Distinct from conventional pipelines that rely solely on the generative priors of generalist MLLMs, our framework explicitly injects Minecraft-specific knowledge (e.g., expert models, knowledge bases, environment feedback) as ground-truth constraints. It employs such knowledge to guide and verify the accuracy of the reasoning content, significantly reducing the hallucinated content in the annotations. This yields \textbf{OptimusM$^4$}, a large-scale, \textbf{M}ulti-\textbf{M}odal \textbf{M}ulti-task \textbf{M}inecraft dataset that provides the rigorous supervision needed to bootstrap the agent's reasoning capabilities.

\textbf{Dual-Router Aligned MoE Architecture for System 1/2 Synergy.} To reconcile the computational conflicts between reflexive action and deliberative reasoning (Challenge \uppercase\expandafter{\romannumeral 2}), we propose the \textbf{Dual-Router Aligned MoE} architecture. As depicted in Figure \ref{fig:fig2}, it structurally enforces the organic coupling of System 1 and System 2 through a novel two-dimensional routing mechanism. Horizontally, we employ a Task Router to assign distinct experts to heterogeneous tasks, achieving orthogonal parameter decoupling to prevent task interference.  Vertically, we introduce a Layer Router that dynamically modulates the inference depth based on the task's cognitive complexity. It constructs a computational ``Fast Path'' with reduced layer activation for latency-sensitive System 1 actions, while preserving a ``Deep Path'' utilizing the full network depth for complex System 2 reasoning. This design breaks the redundancy of static models, enabling efficient, conflict-free coexistence of dual cognitive processes.

\begin{figure*}[htbp]
    \centering
    \includegraphics[width=1\textwidth]{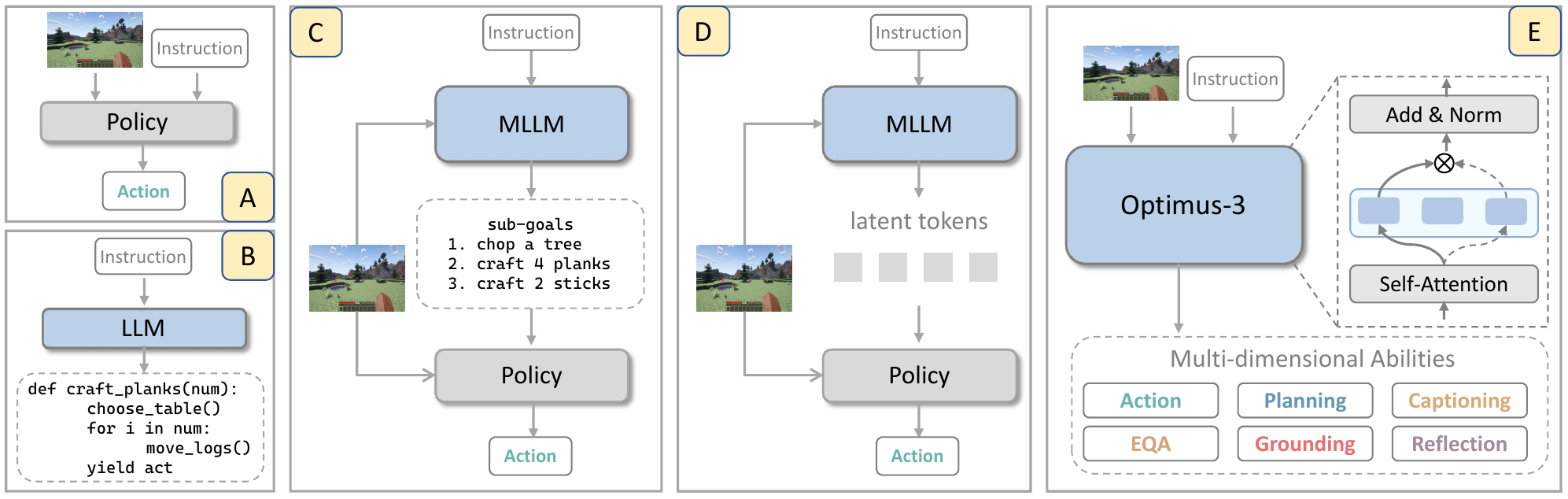}
    \caption{Different agent framework in Minecraft. \textbf{(A)} Goal-conditioned policy which based on Transformer-XL architecture. \textbf{(B)} Function calling, which employs LLM to generate executable functions. \textbf{(C)} (M)LLM as the high-level planner, which then employs a goal-conditioned policy to generate low-level actions. \textbf{(D)} MLLM generates latent tokens that serve as conditioning inputs for the policy. \textbf{(E)} End-to-end MoE architecture (Ours) which endowed with multi-dimensional capabilities.}
    \label{fig:fig3}
\end{figure*}

\textbf{Dual-Granularity Reasoning-Aware Policy Optimization for System 2.} To endow agents with rigorous System 2 capabilities, we propose a \textbf{D}ual-\textbf{G}ranularity \textbf{R}easoning-Aware \textbf{P}olicy \textbf{O}ptimization (DGRPO) algorithm. Unlike conventional RL that relies on sparse outcome signals, DGRPO establishes a novel Process-Outcome Co-Supervision paradigm to mitigate inconsistencies between reasoning process and answers. Central to this framework is our Dual-Granularity Dense Reward mechanism, designed to address specific reasoning failures: (i) The Dependency-Aware Synthesis Reward incorporates domain knowledge graphs to enforce topological consistency, ensuring the agent's thought process strictly adheres to hierarchical logic; (ii) The Hallucination-Aware Consistency Reward acts as a fine-grained perceptual verifier, explicitly penalizing non-existent entities within the reasoning trace to align high-level thought with low-level visual evidence. By providing dense feedback on reasoning process, DGRPO compels the agent to ``think-before-act,'' significantly enhancing robustness and generalization in open-ended dynamic environments.

We conducted comprehensive evaluations in the open-world environment of Minecraft. Experimental results show that Optimus-3 outperforms both general MLLMs and previous SOTA agents in Minecraft across a wide range of tasks. Compared to previous SOTA, Optimus-3 achieves improvements of 21$\%$ on \textit{Planning}, 3$\%$ on \textit{Long-Horizon Action}, 66$\%$ on \textit{Captioning}, 76\% on \textit{Embodied QA}, 3.4$\times$ on \textit{Grounding}, and 18$\%$ on \textit{Reflection}, respectively. Notably, on \textit{Open-Ended} tasks, Optimus-3 achieves an average success rate of 60$\%$, whereas existing agents almost never succeed due to their lack of multi-dimensional capabilities. In summary, our contributions are as follows:
\begin{itemize}
    \item We propose Optimus-3, the first generalist Minecraft agent that organically integrates System 1 action loops with System 2 reasoning capabilities (planning, perception, reflection) within a unified framework. 
    \item We propose a knowledge-enhanced data generation pipeline that synthesizes high-quality System 2 reasoning traces from interaction trajectories. Based on it, we release Multi-Modal, Multi-task Minecraft dataset \textbf{OptimusM$^4$} for the community. 
    \item We design the Dual-Router Aligned MoE architecture, which resolves computational conflicts by decoupling task parameters horizontally and adapting reasoning depth vertically, enabling efficient System 1/2 synergy.
    \item We propose a Dual-Granularity Reasoning-Aware Policy Optimization algorithm. It introduces a process-outcome co-supervision mechanism to align reasoning chains with visual evidence, enabling robust decision-making under dense feedback.
\end{itemize}

\section{Related Work}

\noindent\textbf{Minecraft Agents}. The existing Minecraft agent frameworks are illustrated in Figure \ref{fig:fig3}. Early work \cite{vpt,fan2022minedojo,lifshitz2024steve,cai2023groot,deng2025open} build goal-conditioned policies based on Transformer-XL architecture, refer to Figure \ref{fig:fig3} \textbf{(A)}. Such policies have limited capability in instruction-following and long-horizon planning, and thus can only execute a few simple, atomic tasks. Figure \ref{fig:fig3} \textbf{(B)} show that several works \cite{wang2023voyager,liu2024rl,liu2024odyssey,li2024auto,li2024larm,zhou2024wall} leverage Large Language Models (LLM) to generate executable code, enabling agent-environment interaction through function calling mechanisms. However, this API-call paradigm is fundamentally different from how humans execute low-level actions. To endow agents with human-like coupling between high-level planning and low-level execution, several work \cite{qin2023mp5,wang2023describe,wang2023jarvis,li2024optimus,li2025optimus} employ MLLMs as planners and use the policy as the executor, refer to Figure \ref{fig:fig3} \textbf{(C)}. Moreover, some work \cite{wang2024omnijarvis,cai2025rocket1} leverage MLLMs to generate latent tokens as conditions for the policy, rather than using explicit textual instructions (Figure \ref{fig:fig3} \textbf{(D)}). Despite significant advances, these agents still underperform open-ended tasks depicted in Figure \ref{fig:fig1}, due to their lack of multidimensional competencies. As shown in Figure \ref{fig:fig3} \textbf{(E)}, we develop an end-to-end generalist agent Optimus-3, which is equipped with comprehensive capabilities in Minecraft.

\begin{figure*}[htbp]
    \centering
    \includegraphics[width=0.9\textwidth]{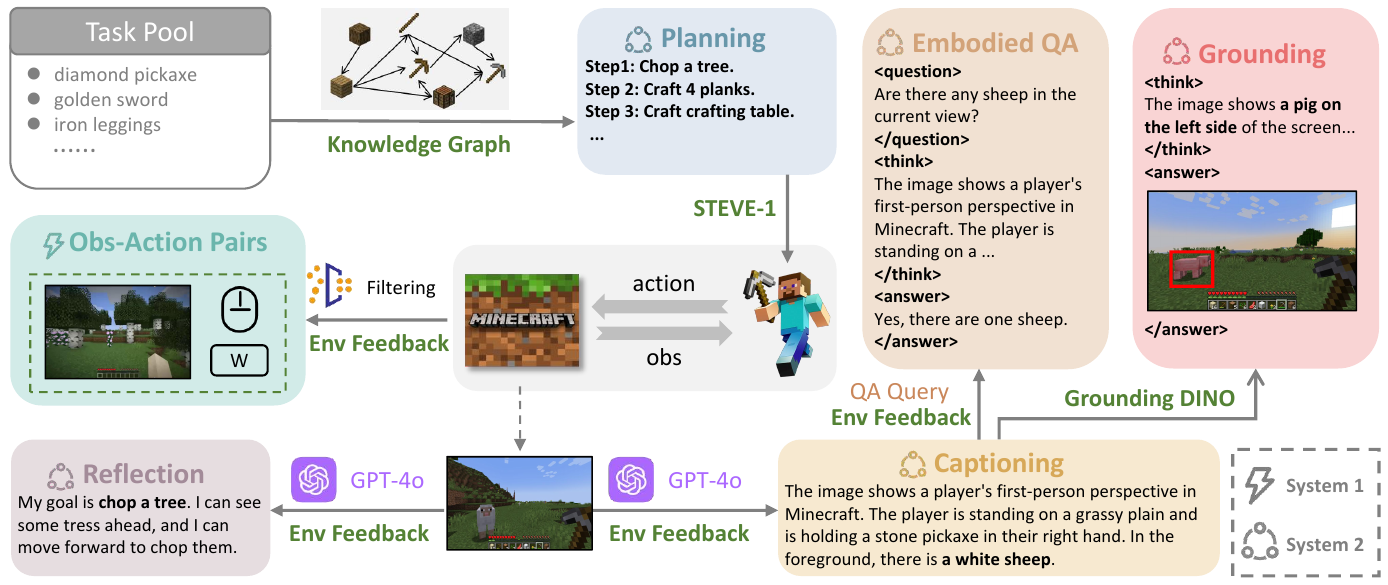}
    \caption{Knowledge-Enhanced Data Generation Pipeline. The  knowledge source is in \textbf{\textcolor{green}{green}}. Given a task pool, we utilize a knowledge graph \cite{li2024optimus} to generate task plans, forming the planning dataset. These plans are then used as instructions for STEVE-1 \cite{lifshitz2024steve}, which interacts with the environment to produce the action dataset. During this process, we randomly sample images and employ expert models \cite{lu2024deepseek,liu2024grounding} with  environmental feedback to generate the captioning, embodied QA, and grounding datasets.}
    \label{fig:fig4}
    
\end{figure*}
\noindent\textbf{Mixture-of-Experts Architectures}. In recent years, Mixture of Experts (MoE) architectures \cite{mu2025comprehensive,moesurvey} have garnered significant attention in the field of Large Language Models (LLMs), owing to their scalability and sparsity in activation \cite{lepikhin2020gshard,fedus2022switch,du2022glam}. To further enhance expert specialization, DeepSeekMoE \cite{dai2401deepseekmoe} proposed more differentiated expert training strategies, significantly boosting downstream task performance, albeit with increased training complexity. Recently, MoE architectures have been incorporated into MLLMs \cite{shen2025mome,li2025uni,lin2024moma,wu2024omni,huang2025modes}, significantly enhancing their generalization capabilities and computational efficiency. Despite these advancements, existing MoE architectures still face notable challenges in load balancing and routing optimization \cite{shen2025mome}. In this paper, we propose a dual-router aligned MoE architecture, in which a task router to achieve orthogonal parameter decoupling, a layer router implements adaptive reasoning depth.

\noindent\textbf{Reinforcement Learning}. Early deep Reinforcement Learning (RL) algorithms \cite{mnih2015dqn,lillicrap2015ddpg,schulman2015trpo} established the foundations of learning control policies from high-dimensional feedback but suffered from instability and limited scalability to large models. Proximal Policy Optimization (PPO) \cite{schulman2017ppo} has since become the mainstream on-policy algorithm for continuous control and LLM fine-tuning, thanks to its clipped surrogate objective that stabilizes policy updates, yet it remains sample-inefficient and sensitive to reward shaping. Building on PPO, Reinforcement Learning from Human Feedback (RLHF) \cite{christiano2017rlhf,ouyang2022instructgpt} trains a reward model from human preferences and then applies PPO-style optimization to align models with nuanced human objectives, but introduces substantial annotation cost and training instability. To simplify this pipeline, Direct Preference Optimization (DPO) \cite{rafailov2023dpo} removes the explicit RL loop by recasting preference alignment as a supervised objective over paired responses, yielding more stable and implementation-friendly training. More recently, Group Relative Policy Optimization (GRPO) \cite{shao2024deepseekmath} replaces the critic in PPO with group-wise relative baselines, significantly reducing memory and computation for large-scale LLMs while preserving on-policy updates. DeepSeek-R1 \cite{guo2025deepseek} enhances the reasoning ability of LLM via GRPO, yet its reward design remains relatively coarse, lacking fine-grained supervision for thinking process and final answers. In this paper, we propose Multimodal Reasoning-Augmented Reinforcement Learning, which assigns task-specific fine-grained rewards to heterogeneous tasks and leverages multimodal reasoning to guide the model to focus on informative visual cues.

\section{Optimus-3}
In this section, we introduce the framework of Optimus-3, designed to organically integrate System 1 action loops with System 2 reasoning capabilities. First, we introduce the \textbf{Knowledge-Enhanced Data Generation Pipeline} (Sec. \ref{pipeline}), which synthesizes high-quality System 2 reasoning traces from raw System 1 trajectories. Then, we detail the \textbf{Dual-Router Aligned MoE Architecture} (Sec. \ref{moe}), specifically engineered to reconcile the computational conflicts between fast, reflexive actions and slow, deliberative reasoning. Subsequently, in Sec \ref{rl}, we elaborate on the \textbf{Multimodal Reasoning-Augmented RL}, a training paradigm designed to activate the agent's System 2 capabilities. Finally, Sec \ref{training_strategy} outlines our multi-stage training strategy.

\subsection{Knowledge-Enhanced Data Generation Pipeline}
\label{pipeline}
\begin{figure*}[htbp]
    \centering
    \includegraphics[width=0.92\textwidth]{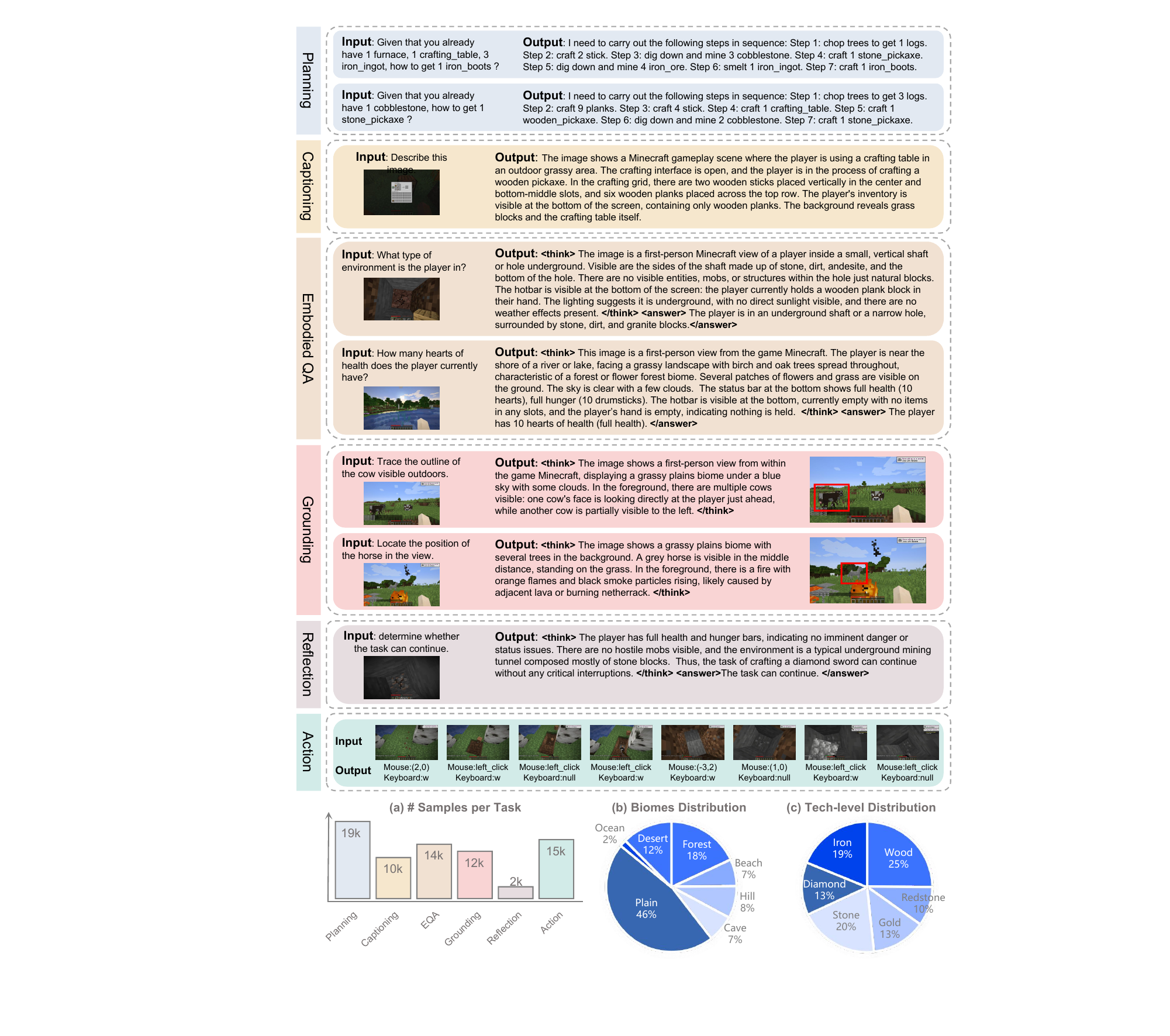}
    \caption{Data visualization and statistics of the OptimusM$^4$ dataset. Top: Representative samples across the Planning, Captioning, Embodied QA, Grounding, Reflection, and Action tasks. Bottom: Statistical overview, detailing sample counts, biome distribution, and tech-level distribution for Action data.}
    \label{fig:fig5}
    
\end{figure*}
Currently, while raw gameplay data (System 1 action trajectories) is abundant, datasets capturing the System 2 reasoning traces are scarce. Relying on general-purpose MLLMs for automated annotation often fails, as they lack the domain-specific knowledge of Minecraft, leading to unfeasible plans or hallucinated responses. For instance, most of generic MLLMs \cite{gpt4,lu2024deepseek,dong2024internlm,team2024gemini} lack an understanding of Minecraft crafting recipes (object synthesis relationships), rendering them incapable of generating feasible plans. Furthermore, these models frequently exhibit hallucinations \cite{bai2024hallucination} in vision-centric tasks, such as dense captioning, Embodied Question Answering (EQA), and visual grounding. To address this, we propose a knowledge-enhanced automated data generation pipeline. It injects external knowledge (e.g., domain knowledge graphs, expert models, environmental feedback) as ground-truth constraints to guide and verify the generation process.

As depicted in Figure \ref{fig:fig4}, we first construct a task pool by collecting general items from Minecraft Wiki. For each rollout, we utilize a domain knowledge graph \cite{li2024optimus} to enforce the topological correctness of crafting paths, ensuring the generated plan is feasible. We then employ an expert policy \cite{lifshitz2024steve} to sequentially execute the planned sub-goals. Trajectories verified as successful via environmental feedback signals are archived as action data (observation-action pairs). During execution, visual frames are sampled at a fixed frequency. Subsequently, we utilize environmental feedback (including agent state, inventory, and surrounding objects) as ground-truth for GPT-4o \cite{gpt4} to generate detailed captions. These captions serve as the thinking process for System 2 tasks, enabling the agent to reason based on  visual observations. The annotation process of System 2 tasks is as follows:
\begin{itemize}
    \item \textbf{Embodied QA:} We employ DeepSeek-VL2 \cite{lu2024deepseek} to generate question-answer pairs derived from the generated captions, ensuring captions and QA pairs are factually aligned with the environment feedback.
    \item \textbf{Visual Grounding:} We incorporate Grounding DINO \cite{liu2023grounding} as a vision expert to annotate objects, providing precise bounding boxes that general MLLMs often miss.
    \item \textbf{Reflection:} We employ GPT-4o \cite{gpt4} to annotate execution status based on environment feedback, creating traces of self-correction.
\end{itemize}
We utilize these knowledge sources to filter out hallucinations, and construct Multi-Modal Multi-task Minecraft dataset \textbf{OptimusM$^4$}. It provides the high-fidelity System 2 supervision required to bootstrap the agent's reasoning capabilities. Data visualizations and detailed statistics are presented in Figure \ref{fig:fig5}.

\subsection{Dual-Router Aligned MoE Architecture}
\label{moe}
Unified modeling of System 1 and System 2 introduces a fundamental architectural dilemma: these two modes impose distinct computational requirements. System 1 (Action) demands high-frequency, latency-sensitive processing of local contexts, whereas System 2 (Planning, Reflection) requires deep, computation-intensive reasoning over long horizons. Standard dense architectures enforce a uniform depth, making them either too slow for real-time control or too shallow for complex reasoning. Furthermore, co-training these heterogeneous tasks often leads to gradient interference and negative transfer \cite{moesurvey,huang2025modes}. To resolve these conflicts, we propose the \textbf{Dual-Router Aligned MoE}, a task-aware architecture that structurally enforces System 1/2 coupling through a two-dimensional routing mechanism (Figure \ref{fig:fig2}).

\subsubsection{Horizontal Routing: Orthogonal Parameter Decoupling}
To prevent gradient interference between conflicting tasks (e.g., visual grounding vs. action prediction), we employ a semantic Task Router.  Unlike token-level soft routing which can lead to load imbalance \cite{shen2025mome}, it deterministically directs input tokens to topologically distinct parameter spaces based on the task type $\mathcal{T}$.

For the $l$-th layer, we replace the standard Feed-Forward Network (FFN) with a hybrid expert module comprising a Shared Knowledge Expert $E_\text{S}$ and Task-Specific Experts $E_\mathcal{T}$. The output hidden state $\mathbf{h}^l$ is computed as:
\begin{equation}
\label{eq:horizontal_routing}
\mathbf{h}^l = \mathbf{x}^l  \oplus (\underbrace{E_{\text{S}}(\mathbf{x}^l)}_{\text{General Knowledge}} + \underbrace{E_{\mathcal{T}}(\mathbf{x}^l)}_{\text{Task-Specific}}),
\end{equation}
where $\mathbf{x}^l$ denotes the input features (after Attention and Normalization). Here, $E_\text{S}$ captures universal semantic patterns across tasks to facilitate positive transfer, while $E_\mathcal{T}$ is exclusively activated for task $\mathcal{T}$. This orthogonal design ensures that the optimization of System 1 control policies is isolated from System 2 reasoning tasks, maintaining a shared representational foundation without task interference.

\subsubsection{Vertical Routing: System-1/2 Adaptive Reasoning}
Complementing the horizontal separation, we introduce a Layer Router to align the model's reasoning depth with the cognitive complexity of the task. Grounded in the Dual-Process Theory \cite{kahneman2011thinking}, this router implements a \textbf{Dynamic Compute Allocation} mechanism. It recognizes that reflexive tasks (System 1) prioritize inference speed, whereas deliberative tasks (System 2) demand reasoning depth.

Formally, we model the vertical routing as a layer selection problem based on task-specific importance. Let $\mathbf{e}_{\mathcal{T}}$ denote the task embedding. We employ a layer router to project the task representation into a layer-wise importance distribution $\boldsymbol{\alpha} \in \mathbb{R}^L$ via a Softmax function:
\begin{equation}
\label{eq:importance_distribution}
\boldsymbol{\alpha} = \operatorname{Softmax}\left( \operatorname{MLP}(\mathbf{e}_{\mathcal{T}}) \right),
\end{equation}
where $\alpha_l$ quantifies the contribution of the $l$-th layer to the current task. To achieve adaptive inference, we introduce a filtering threshold $\tau$ to dynamically determine the set of active layers $\Phi_{\mathcal{T}}$:
\begin{equation}
\Phi_{\mathcal{T}} = \left\{ l \in \{1, \dots, L\} \mid \alpha_l \ge \tau \right\}.
\end{equation}
During the forward pass, layers with importance scores below the threshold are bypassed via residual connections. The state transition is formulated as:
\begin{equation}
\mathbf{x}^{l+1} = \mathbf{x}^l + \mathbb{I}(l \in \Phi_{\mathcal{T}}) \cdot \left( \text{Attn}(\mathbf{x}^l) + \text{MoEBlock}_l(\mathbf{x}^l) \right),
\end{equation}
where $\mathbb{I}(\cdot)$ is the indicator function. This mechanism naturally induces a cognitive dichotomy:
\begin{itemize}
    \item \textbf{System 2 Mode (Deep Path):} For reasoning-intensive tasks (e.g., Planning), the router yields a high-entropy distribution or uniformly high importance across critical reasoning blocks, resulting in $|\Phi_{\mathcal{T}}| \approx L$ to ensure deep reasoning.
    \item \textbf{System 1 Mode (Fast Path):} For interaction-intensive tasks (e.g., Action), the router assigns high probability mass only to a few essential perceptual and control layers. This filters out redundant intermediate computation (i.e., $|\Phi_{\mathcal{T}}| \ll L$), significantly reducing latency.
\end{itemize}
By selectively skipping calculations for non-essential layers, the Dual-Router Aligned MoE achieves a dynamic equilibrium between computational efficiency and reasoning capability, effectively synthesizing the reflexive nature of System 1 and the deliberative depth of System 2 within a unified framework.

\begin{figure*}[htbp]
    \centering
    \includegraphics[width=0.92\textwidth]{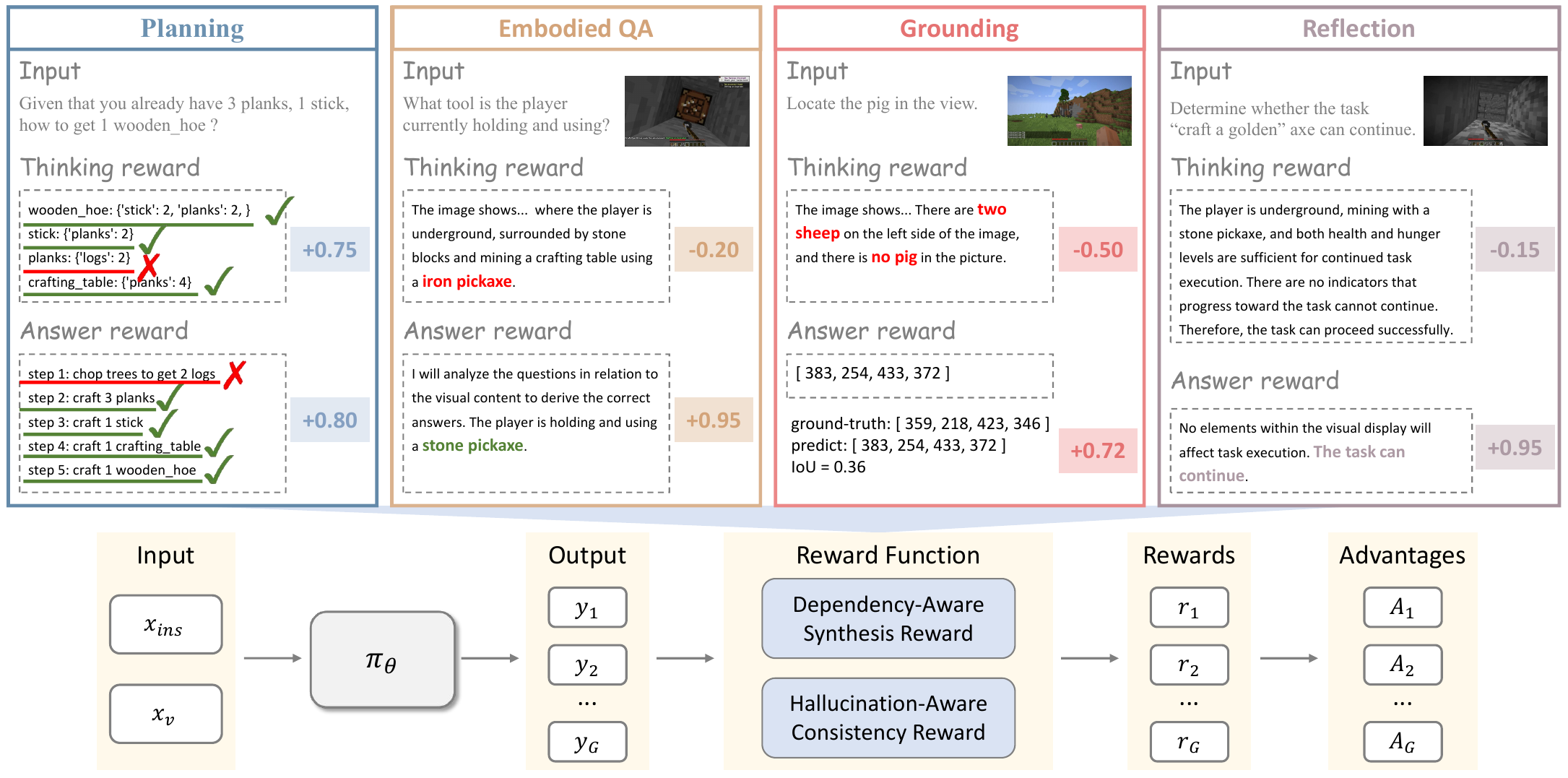}
    \caption{Visualization examples of the task-specific fine-grained reward functions in DGRPO. For the Planning task, we design a Dependency-Aware Synthesis Reward, which treats the item's crafting dependency path as thinking reward and assigns fine-grained step-wise supervision as answer reward. For vision-related tasks, we introduce a Hallucination-Aware Consistency Reward that penalizes hallucinated items in the reasoning process and the final answer.
}
    \label{fig:fig6} 
\end{figure*}

\subsection{Dual-Granularity Reasoning-Aware Policy Optimization}
\label{rl}

In the open-ended and dynamic environment of Minecraft, relying solely on Supervised Fine-Tuning (SFT) is insufficient. While Reinforcement Learning (RL) \cite{shao2024deepseekmath} offers a solution to activate reasoning capabilities through exploration \cite{guo2025deepseek}, it lacks fine-grained supervision of the reasoning process. To this end, we propose the \textbf{Dual-Granularity Reasoning-Aware Policy Optimization (DGRPO)} algorithm. It explicitly activates System 2 capabilities by establishing a Process-Outcome Co-Supervision paradigm, utilizing dense rewards to enforce logical consistency between the intermediate reasoning and the final answer.

\subsubsection{Overview of Framework}
DGRPO consists of two progressive phases: a visual-reasoning cold-start phase and a preference alignment phase via reinforcement learning.

\noindent\textbf{Phase 1: Visual-Reasoning Cold Start.}
To initialize the model with basic reasoning capabilities, we perform fine-tuning using Chain-of-Thought (CoT) \cite{wei2022chain} templates. Unlike standard text-only CoT, our approach compels the model to explicitly describe the visual scene within its reasoning trace, thereby grounding high-level reasoning in low-level visual evidence. The objective is defined as:
\begin{equation}
\mathcal{L}_{\text{SFT}} = - \sum_{t=1}^{T} \log \pi_{\theta} \left( y_t \mid x_v, x_{\text{ins}}, y_{<t} \right), \;\;y = [ y^{(\text{think})}; y^{(\text{ans})} ].
\end{equation}
Here, $x_v$ and $x_{\text{ins}}$ denote visual and instructional inputs, respectively. The output $y$ is decomposed into a reasoning process $y^{(\text{think})}$ and a final answer $y^{(\text{ans})}$. This phase serves to mitigate initial hallucinations and activate the model's reasoning capabilities.

\noindent\textbf{Phase 2: Optimization via DGRPO.}
To further robustify the reasoning process, we employ Group Relative Policy Optimization (GRPO) \cite{shao2024deepseekmath} as the optimization backbone. GRPO is particularly advantageous for large-scale MLLMs as it eliminates the need for a separate value function critic. Instead, it utilizes the group average reward as a baseline, reducing computational overhead while stabilizing training.
Formally, given an input $x$, the policy $\pi_{\theta}$ samples a group of $G$ outputs $\{y_1, y_2, \dots, y_G\}$. The optimization objective is formulated as:
\begin{equation}
\begin{split}
\frac{1}{G} \sum_{i=1}^{G} \frac{1}{\left| y_{i} \right|} \sum_{t=1}^{\left| y_{i} \right|}
\Big\{
\min \Big[
  r_{\theta}\left(x, y_{i}\right) A_{i,t},
  \\
  \operatorname{clip}\big( r_{\theta}\left(x, y_{i}\right), 
  1-\varepsilon, 1+\varepsilon \big) A_{i,t}
\Big] - \lambda D_{KL}
\Big\}
\end{split}
\end{equation}

\begin{equation}
r_{\theta }\left ( x, y_{i}  \right ) = \frac{\pi _{\theta } \left ( y_{i,t} \mid x,y_{i,<t}  \right )  }{\pi _{\theta_{old}} \left ( y_{i,t} \mid x,y_{i,<t}  \right ) } , 
\end{equation}

\begin{equation}
D_{KL} = \frac{\pi _{ref } \left ( y_{i,t} \mid x,y_{i,<t}  \right )  }{\pi _{\theta} \left ( y_{i,t} \mid x,y_{i,<t}  \right ) } - \log\frac{\pi _{ref } \left ( y_{i,t} \mid x,y_{i,<t}  \right )  }{\pi _{\theta} \left ( y_{i,t} \mid x,y_{i,<t}  \right ) }  -1,
\end{equation}
where \( A_{i,t} \) denotes the advantage function, \( \varepsilon \) and \( \lambda  \) are hyperparameters, and \( \pi_{\theta_{old}} \) and \( \pi_{ref} \) represent the old policy and the reference model, respectively. 

\subsubsection{Dual-Granularity Dense Reward Design}
Standard GRPO algorithm relies on sparse outcome rewards (e.g., success/failure), which treat the reasoning process as a black box. Crucially, such answer-only supervision fails to penalize flawed reasoning that accidentally yields the correct response (i.e., spurious correlations). DGRPO addresses this by introducing a \textbf{Process-Outcome Co-Supervision} mechanism. We design task-specific dense rewards that provide fine-grained feedback on both the \textit{reasoning process} ($R_{\text{think}}$) and the \textit{final answer} ($R_{\text{ans}}$). As illustrated in Fig. \ref{fig:fig6}, the reward system is tailored for two primary task categories.

\noindent\textbf{Dependency-Aware Synthesis Reward}.
In long-horizon planning tasks, the core challenge is strictly adhering to the hierarchical synthesis logic of Minecraft. We propose the Dependency-Aware Synthesis Reward ($R_{\text{DAS}}$) to enforce topological consistency with the domain knowledge graph.
Let $\mathcal{G}_{\text{syn}}$ be the ground-truth synthesis dependency graph. The total reward $R_{\text{DAS}}$ is a weighted sum of the thinking reward $R_\text{think}$ and plan accuracy reward $R_\text{ans}$:
\begin{equation}
R_{\text{DAS}}(y) = \underbrace{\frac{1}{M} \sum_{m=1}^{M} \mathbb{I}(y^{(\text{think})}_m \in \mathcal{G}_{\text{syn}})}_{R_{\text{think}}} + \underbrace{\frac{1}{N} \sum_{n=1}^{N} \mathbb{I}(y^{(\text{ans})}_n = y^{(\text{gold})}_n)}_{R_{\text{ans}}},
\end{equation}
where $\mathbb{I}(\cdot)$ is the indicator function, and $M$ and $N$ represent the number of reasoning steps and plan steps, respectively. $R_{\text{thinking}}$ encourages the model to explicitly verify material prerequisites in its thought chain, while $R_{\text{answer}}$ ensures the final plan sequence matches the optimal path.

\noindent\textbf{Hallucination-Aware Consistency Reward}.
For tasks requiring precise visual grounding (Embodied QA, Reflection, Grounding), the primary failure mode is perceptual hallucination \cite{bai2024hallucination}. We introduce the Hallucination-Aware Consistency Reward ($R_{\text{HAC}}$) to penalize the generation of non-existent entities during reasoning.
Let $\mathcal{S}_{\text{gold}}$ denote the set of ground-truth objects present in the scene. The reward function is defined as:
\begin{equation}
R_{\text{HAC}}(y) = \underbrace{\frac{1}{N} \sum_{n=1}^{N} \mathbb{I}(y^{(\text{think})}_n \in \mathcal{S}_{\text{gold}})}_{R_{\text{think}}} + \underbrace{r_{\text{task}}(y^{(\text{ans})}, y^{(\text{gold})})}_{R_{\text{ans}}}.
\end{equation}
Here, $R_\text{think}$ explicitly measures perceptual precision (analogous to $1 - \texttt{CHAIR}$ \cite{rohrbach2018object}), forcing the agent to ground its reasoning in visual perception to answer questions.
The answer reward $R_\text{ans}$ adapts to the specific task format:

\begin{itemize}
    \item \textbf{Embodied QA}: We utilize ROUGE-L F1 score to measure the accuracy of the agent's responses.
    \item \textbf{Reflection}: We utilize binary success indicators to measure accuracy of the agent's decisions.
    \item \textbf{Visual Grounding:} We implement a strict \textbf{IoU-Density Reward} to penalize imprecise bounding boxes. Given the Intersection-over-Union (IoU) score $u$ between predicted and ground-truth boxes:
    \begin{equation}
    R_{\text{ans}} = 
    \begin{cases}
    1, & \text{if } u \ge \alpha \\
    \eta u, & \text{if } \beta \le u < \alpha \\
    0, & \text{if } u < \beta
    \end{cases}
    \end{equation}
    Here, $\alpha$ and $\beta$ are hyperparameters in the range $(0, 1)$, and $\eta$ is a weighting coefficient.
\end{itemize}

\begin{figure}[htbp]
\centering
\includegraphics[width=0.45\textwidth]{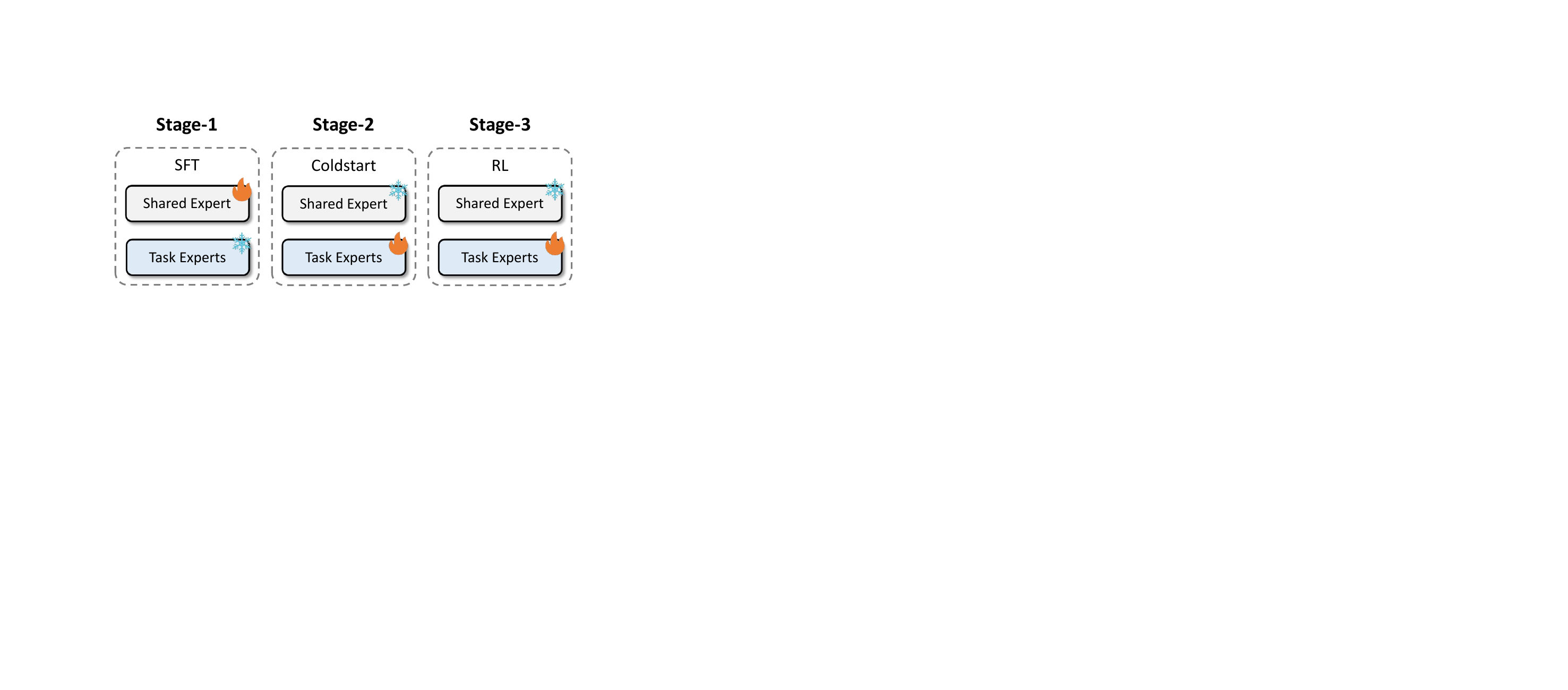}
\caption{The stage-wise training strategy. \textbf{Stage-1}: We train the shared knowledge expert via Supervised Fine-Tuning (SFT). \textbf{Stage-2}: We train the task experts with multimodal reasoning fine-tuning (Coldstart). \textbf{Stage-3}: We further train the task experts with Reinforcement Learning (RL).}
\label{fig:fig7}
\end{figure}
Additionally, we add a Format Reward with weight $\lambda_f$ to each task, requiring the agent's response to follow the format $<\texttt{think}>...</\texttt{think}><\texttt{answer}>...</\texttt{answer}>$.

\subsection{Training Strategy}
\label{training_strategy}
To ensure stable convergence, we adopt a three-stage training strategy (as illustrated in Fig. \ref{fig:fig7}) for Optimus-3.

\noindent\textbf{Stage-1}: In this stage, we initialize the shared knowledge expert from a dense model, then train it via supervised fine-tuning. The shared knowledge expert is trained across all tasks to capture common semantic representations. 

\noindent\textbf{Stage-2}: In this stage, we initialize the task experts from shared expert, then freeze the shared knowledge expert. The task experts are trained via the fine-tuning (coldstart) to activate multimodal reasoning capabilities. While the Action expert is trained via imitation learning \cite{li2025optimus}.

\noindent\textbf{Stage-3}: We refine the System 2 experts (Planning, Perception, Reflection) via the DGRPO algorithm. By optimizing against the proposed dual-granularity rewards, the agent learns to generate reasoning chains that are both logically sound and visually faithful.

\section{Experiments}

\subsection{Experiments Setting}

\noindent\textbf{Environment}. Following the standard evaluation protocol adopted in prior studies \cite{lifshitz2024steve,wang2023describe,wang2023jarvis,qin2023mp5,li2024optimus}, we conduct experiments in the open-world environment of Minecraft on the MineRL \cite{guss2019minerl} platform. It is an archetypal open-world environment that offers rich diversity in resources and biomes, thereby requiring agents to operate under long-horizon and highly variable conditions. In MineRL, the agent receives instructions and observations then produces mouse/keyboard control actions at 20 Hz. For each episode, the agent is spawned without any initial equipment and is initialized at random biomes and positions, leading to substantial variation in environmental dynamics and task difficulty. Therefore, Minecraft serves as a suitable and challenging testbed for evaluating open-world agents.
\begin{table}[ht]
\centering
\caption{Hyperparameter setting for each training phase.}
\label{tab:hyper}
\resizebox{0.45\textwidth}{!}{%
\renewcommand\arraystretch{1.2}
\begin{tabular}{lccc}
\toprule[1.2pt]
Hyperparameter & Stage-1 & Stage-2 & Stage-3 \\ \hline
Optimizer      & AdamW   & AdamW      & AdamW       \\
Learning Rate  & 5.0e-5   & 3.0e-5   & 1.0e-6     \\
Epochs         & 2     & 2       & 20           \\
Batch Size      & 12    & 8    & 16        \\
Gradient Accumulation & 16 & 16 & 8 \\
Warm Up Ratio & 0.25 & 0.25 & - \\
Max Pixels & 234416 & 234416 & 234416 \\
Num Generations & - & - & 5 \\
Max Prompt Length & 2048 & 2048 & 2048 \\
\bottomrule[1.2pt]
\end{tabular}
}
\end{table}

\noindent\textbf{Implementation details}. We initialize Optimus-3 with the weights of Qwen2.5-VL-7B \cite{bai2025qwen2}. It comprises a large language model (LLM) and a ViT-based \cite{dosovitskiy2020vit} visual encoder. Then we adapt it into MoE architecture, comprising one shared knowledge expert and five task-specific experts dedicated to planning, perception, action, grounding, and reflection. Subsequently, we implement the Task Router by fine-tuning a lightweight Sentence-BERT \cite{reimers2019sentence} model to classify instructions into the predefined task set. For Action tasks, following prior work \cite{li2025optimus}, we employ  VPT \cite{vpt} as action head to generate low-level actions. We collect 230k samples for training stage-1, 58k samples for the training stage-2, and 5k samples for training stage-3. These datasets are sourced from our proposed OptimusM$^4$ as well as previous works \cite{vpt,li2025optimus,qin2023mp5}. All experiments were conducted on 8$\times$ NVIDIA L40 GPUs. The hyperparameter setting are shown in Table \ref{tab:hyper}.

\begin{table*}[t]
\scriptsize
\centering
\caption{Evaluation Results of Optimus-3 on MineSys2 benchmark. \#Params denotes activated parameters. SFT and RL refer to supervised fine-tuning and reinforcement learning, respectively.}
\label{tb:main_text}
\renewcommand\arraystretch{1.1}
\resizebox{0.8\textwidth}{!}{%

\begin{tabular}{lcccccc}
\toprule[1.2pt]
\multirow{2}{*}{Model} & \multirow{2}{*}{\#Params} & Planning & Captioning & EQA & Grounding & Reflection \\
      &          & \texttt{Acc}      & \texttt{Score}      & \texttt{Score} & \texttt{IoU@0.5}  & \texttt{Acc} \\
\hline
\multicolumn{7}{c}{\textit{Generalist Multimodal Large Language Model}} \\
\hline
GPT-4o   \cite{gpt4}       &   \multicolumn{1}{c}{-}  &  0.20   &   0.46  & 0.33   &  -  & 0.31   \\
Gemini-1.5-pro  \cite{team2024gemini}      &    \multicolumn{1}{c}{-} &   0.19   & 0.33    & 0.33   & -   &  0.34    \\
DeepSeek-VL2 \cite{lu2024deepseek}        &   4B  &   0.07   &  0.49  &   0.51  &  0.29  &  0.47  \\
Qwen2.5-VL \cite{bai2025qwen2}          &   3B  &  0.03    & 0.37    & 0.40   &  0.58  & 0.47   \\
Qwen2.5-VL \cite{bai2025qwen2}          &   7B  &   0.05   &  0.47  &   0.46  &  0.18  &  0.56  \\
Qwen2.5-VL \cite{bai2025qwen2}          &   32B  &  0.07    & 0.51    & 0.49   &  0.34  & 0.53   \\
Qwen2.5-VL \cite{bai2025qwen2}          &   72B  &   0.09   &  0.52  &   0.54  &  0.32  &  0.53  \\

\hline
\multicolumn{7}{c}{\textit{Post-trained Multimodal Large Language Model}} \\
\hline
Qwen2.5-VL-SFT \cite{bai2025qwen2}      &   3B    &   0.76  & 0.64  &  0.69 & 0.69   & 0.47     \\
Qwen2.5-VL-RL  \cite{bai2025qwen2}      &   3B   &     0.74 & 0.65    &  0.70  &  0.71  &  0.48    \\
Qwen2.5-VL-SFT  \cite{bai2025qwen2}     &   7B    &   0.79   & 0.68   &   0.71 &   0.52   & 0.53      \\
Qwen2.5-VL-RL  \cite{bai2025qwen2}      &   7B   &   0.76   & 0.71    &   0.68 &  0.79  & 0.56     \\ 

\hline
\rowcolor[HTML]{E7EEFE}
Optimus-3 $_{[token\;routing]}$            &    6.8B &  0.88   &   0.66  & 0.77   &  0.75  &  0.58    \\
\rowcolor[HTML]{E7EEFE}
Optimus-3 $_{[task\;routing]}$            &    6.8B &   \textbf{0.94}   &   \textbf{0.78}  & \textbf{0.81}   & \textbf{0.80}  &    \textbf{0.66}   \\
\bottomrule[1.2pt]
\end{tabular}
}
\end{table*}


\begin{table*}[htbp]
\centering
\caption{Main Result of Optimus-3 on Long-Horizon Benchmark. We report \texttt{SR} on each task group. ${\dagger}$ denotes we employ a hierarchical architecture agent from prior work \cite{li2025optimus}, where an MLLM serves as the planner, followed by STEVE-1 \cite{lifshitz2024steve} acting as the policy to sequentially generate actions. ${*}$ represents reproduced results under the same settings as other baselines. Optimus-3-Action denotes Optimus-3 trained only on action trajectories.}
\label{tb:main_horizon}
\scriptsize
\renewcommand\arraystretch{1.2}
\resizebox{1\textwidth}{!}{%
\begin{tabular}{lccccccc}
\toprule[1.2pt]
Method       & Wood \includegraphics[width=0.3cm]{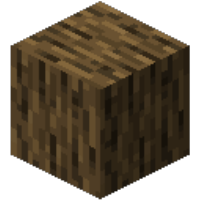} & Stone \includegraphics[width=0.3cm]{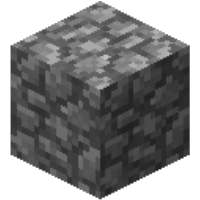} & Iron \includegraphics[width=0.3cm]{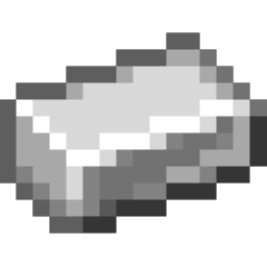} & Gold \includegraphics[width=0.3cm]{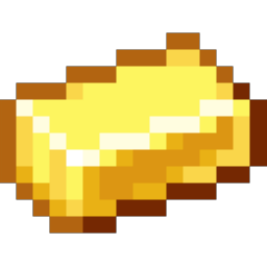} & Diamond \includegraphics[width=0.3cm]{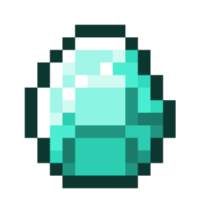} & RedStone \includegraphics[width=0.3cm]{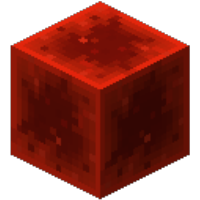} & Armor \includegraphics[width=0.3cm]{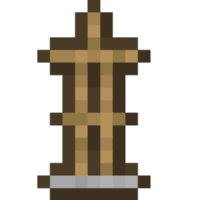}  \\
\hline
\multicolumn{8}{c}{\textit{Multimodal Large Language Model$^{\dagger}$}} \\
\hline
 GPT-3.5 \cite{gpt3}      &    0.40$\scriptstyle\pm\,0.15$  &   0.20$\scriptstyle\pm\,0.13$    &   0.00$\scriptstyle\pm\,0.00$  &   0.00$\scriptstyle\pm\,0.00$   &     0.00$\scriptstyle\pm\,0.00$    &       0.00$\scriptstyle\pm\,0.00$   &    0.00$\scriptstyle\pm\,0.00$           \\
GPT-4o \cite{gpt4}    &    0.47$\scriptstyle\pm\,0.23$   &   0.23$\scriptstyle\pm\,0.09$    &   0.05$\scriptstyle\pm\,0.04$  &   0.00$\scriptstyle\pm\,0.00$   &     0.00$\scriptstyle\pm\,0.00$    &       0.00$\scriptstyle\pm\,0.00$   &    0.00$\scriptstyle\pm\,0.00$           \\
Gemini-1.5-pro \cite{team2024gemini}   &  0.41$\scriptstyle\pm\,0.14$    &  0.21$\scriptstyle\pm\,0.10$    & 0.03$\scriptstyle\pm\,0.02$    &   0.00$\scriptstyle\pm\,0.00$   &     0.00$\scriptstyle\pm\,0.00$    &       0.00$\scriptstyle\pm\,0.00$   &    0.00$\scriptstyle\pm\,0.00$           \\
Qwen2.5-VL \cite{bai2025qwen2}     &  0.28$\scriptstyle\pm\,0.15$    &   0.06$\scriptstyle\pm\,0.03$    &  0.00$\scriptstyle\pm\,0.00$   &   0.00$\scriptstyle\pm\,0.00$  &     0.00$\scriptstyle\pm\,0.00$    &       0.00$\scriptstyle\pm\,0.00$   &    0.00$\scriptstyle\pm\,0.00$           \\
Qwen2.5-VL-SFT \cite{bai2025qwen2}    &  0.76$\scriptstyle\pm\,0.11$    & 0.36$\scriptstyle\pm\,0.07$      &   0.11$\scriptstyle\pm\,0.05$  &  0.00$\scriptstyle\pm\,0.00$    &   0.00$\scriptstyle\pm\,0.00$     &     0.00$\scriptstyle\pm\,0.00$    &   0.00$\scriptstyle\pm\,0.00$          \\
\hline
\multicolumn{8}{c}{\textit{Goal-conditioned Policy in Minecraft}} \\
\hline
VPT \cite{vpt}   \texttt{[NeurIPS'22]}   &  0.18$\scriptstyle\pm\,0.15$   &  0.07$\scriptstyle\pm\,0.05$    &   0.00$\scriptstyle\pm\,0.00$   &  0.00$\scriptstyle\pm\,0.00$    &   0.01$\scriptstyle\pm\,0.01$      &   0.00$\scriptstyle\pm\,0.00$       &   0.00$\scriptstyle\pm\,0.00$           \\
GROOT \cite{cai2023groot}   \texttt{[ICLR'24]}      &  0.34$\scriptstyle\pm\,0.17$   &  0.17$\scriptstyle\pm\,0.10$    &   0.08$\scriptstyle\pm\,0.05$   &  0.01$\scriptstyle\pm\,0.01$    &   0.01$\scriptstyle\pm\,0.01$      &   0.03$\scriptstyle\pm\,0.02$       &   0.04$\scriptstyle\pm\,0.02$           \\
MineCLIP \cite{fan2022minedojo}  \texttt{[NeurIPS'22]}      &  0.23$\scriptstyle\pm\,0.16$   &  0.12$\scriptstyle\pm\,0.08$    &   0.06$\scriptstyle\pm\,0.05$   &  0.00$\scriptstyle\pm\,0.00$    &   0.00$\scriptstyle\pm\,0.00$      &   0.00$\scriptstyle\pm\,0.00$       &   0.02$\scriptstyle\pm\,0.02$           \\
STEVE-1 \cite{lifshitz2024steve}    \texttt{[NeurIPS'23]}     &  0.45$\scriptstyle\pm\,0.22$   &  0.22$\scriptstyle\pm\,0.19$    &   0.08$\scriptstyle\pm\,0.06$   &  0.00$\scriptstyle\pm\,0.00$    &   0.05$\scriptstyle\pm\,0.03$      &   0.00$\scriptstyle\pm\,0.00$       &   0.07$\scriptstyle\pm\,0.05$           \\

\hline
\multicolumn{8}{c}{\textit{Agents in Minecraft}} \\
\hline
Voyager$^{*}$ \cite{wang2023describe}    \texttt{[NeurIPS'23]}    &  0.87$\scriptstyle\pm\,0.25$   &  0.32$\scriptstyle\pm\,0.15$    &   0.08$\scriptstyle\pm\,0.06$   &  0.02$\scriptstyle\pm\,0.02$    &   0.01$\scriptstyle\pm\,0.01$      &   0.00$\scriptstyle\pm\,0.00$       &   0.14$\scriptstyle\pm\,0.09$           \\
DEPS \cite{wang2023describe}    \texttt{[NeurIPS'23]}     &  0.77$\scriptstyle\pm\,0.13$   &  0.48$\scriptstyle\pm\,0.09$    &   0.16$\scriptstyle\pm\,0.08$   &  0.00$\scriptstyle\pm\,0.00$    &   0.01$\scriptstyle\pm\,0.01$      &   0.00$\scriptstyle\pm\,0.00$       &   0.10$\scriptstyle\pm\,0.18$           \\
MP5$^{*}$ \cite{qin2023mp5}    \texttt{[CVPR'24]}     &  0.89$\scriptstyle\pm\,0.23$   &  0.73$\scriptstyle\pm\,0.21$    &   0.43$\scriptstyle\pm\,0.18$   &  0.10$\scriptstyle\pm\,0.08$    &   0.09$\scriptstyle\pm\,0.08$      &   0.17$\scriptstyle\pm\,0.08$       &   0.19$\scriptstyle\pm\,0.18$           \\
JARVIS-1 \cite{wang2023jarvis}    \texttt{[TPAMI'25]}  &  0.93$\scriptstyle\pm\,0.14$   &    0.89$\scriptstyle\pm\,0.07$   &  0.36$\scriptstyle\pm\,0.06$    &  0.07$\scriptstyle\pm\,0.03$    &    0.08$\scriptstyle\pm\,0.03$    &     0.16$\scriptstyle\pm\,0.07$     &   0.15$\scriptstyle\pm\,0.19$            \\
Optimus-1 \cite{li2024optimus}  \texttt{[NeurIPS'24]}  &  0.98$\scriptstyle\pm\,0.02$   & 0.92$\scriptstyle\pm\,0.04$    & 0.46$\scriptstyle\pm\,0.09$    &   0.08$\scriptstyle\pm\,0.05$  &     0.11$\scriptstyle\pm\,0.05$   &   0.25$\scriptstyle\pm\,0.03$    &   0.19$\scriptstyle\pm\,0.22$          \\ 
Optimus-2 \cite{li2025optimus}   \texttt{[CVPR'25]}   & 0.99$\scriptstyle\pm\,0.02$   & 0.93$\scriptstyle\pm\,0.04$    & 0.53$\scriptstyle\pm\,0.03$   &   0.09$\scriptstyle\pm\,0.01$  &   0.13$\scriptstyle\pm\,0.02$    &    0.28$\scriptstyle\pm\,0.03$   &   0.21$\scriptstyle\pm\,0.19$   \\ 
\hline
\rowcolor[HTML]{E7EEFE}
Optimus-3-Action      &    0.93$\scriptstyle\pm\,0.03$ & 0.87$\scriptstyle\pm\,0.05$      &  0.49$\scriptstyle\pm\,0.04$   &  0.03$\scriptstyle\pm\,0.04$  &  0.03$\scriptstyle\pm\,0.02$      &   0.09$\scriptstyle\pm\,0.05$     &   0.15$\scriptstyle\pm\,0.22$  \\
\rowcolor[HTML]{E7EEFE}
Optimus-3      &    \textbf{0.99$\scriptstyle\pm\,0.01$} & \textbf{0.95$\scriptstyle\pm\,0.02$}      &  \textbf{0.55$\scriptstyle\pm\,0.03$}   &  \textbf{0.10$\scriptstyle\pm\,0.02$}  &  \textbf{0.15$\scriptstyle\pm\,0.02$}      &   \textbf{0.29$\scriptstyle\pm\,0.02$}     &   \textbf{0.23$\scriptstyle\pm\,0.16$}  \\

\bottomrule[1.2pt]
\end{tabular}
\vspace{-5pt}
}

\end{table*}


\subsection{Evaluation on System 2 Tasks}
\noindent\textbf{Evaluation Tasks \& Metrics}. To comprehensively evaluate the multi-dimensional cognitive abilities of Optimus-3, we construct a MineSys2 Benchmark comprising five categories: \textit{Planning}, \textit{Captioning}, \textit{Embodied QA}, \textit{Grounding}, and \textit{Reflection}. For \textit{Planning} and \textit{Reflection} tasks, evaluation samples are 103 and 64, respectively. We employ average accuracy (\texttt{Acc}) as the evaluation metric. For the \textit{Captioning} and \textit{Embodied QA} tasks, the evaluation includes 134 and 400 samples, respectively. We adopt an LLM-as-Judge \cite{li2024generation} approach, employing GPT-4 \cite{gpt4} to assign a score from 1 to 10 for each sample. The average score is then normalized to a value between 0 and 1. For the \textit{Grounding} tasks, we construct 500 evaluation samples, and use \texttt{IOU@0.5} \cite{chen2023advancing} as the metric.

\begin{table*}[t]
\centering

\caption{Performance comparison on Wooden Pickaxe \includegraphics[width=0.3cm]{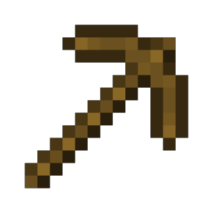}, Stone Sword \includegraphics[width=0.3cm]{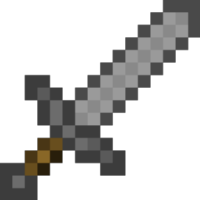}, Iron Ingot \includegraphics[width=0.3cm]{figures/logo/iron_ingot.pdf}, Golden Shovel \includegraphics[width=0.3cm]{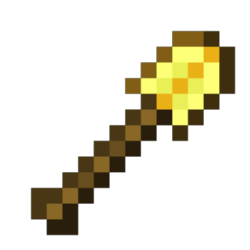}, and Diamond Sword \includegraphics[width=0.3cm]{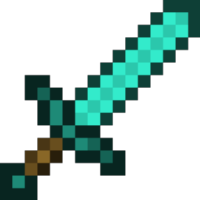}. We report Completion Rate (\texttt{CR}) and Success Rate (\texttt{SR}) for each task. ${\dagger}$ denotes we employ a hierarchical architecture agent from prior work \cite{li2025optimus}. ${*}$ represents reproduced results under the same settings. }
\label{tb:main_open}
\renewcommand\arraystretch{1.2}
 
\resizebox{1\textwidth}{!}{%

\begin{tabular}{lcccccccccccc}
\toprule[1.2pt]
\multirow{2}{*}{Model} & \multicolumn{2}{c}{Wooden Pickaxe \includegraphics[width=0.3cm]{figures/logo/wooden_pickaxe.pdf}} & \multicolumn{2}{c}{Stone Sword \includegraphics[width=0.3cm]{figures/logo/stone_sword.pdf}} & \multicolumn{2}{c}{Iron Ingot \includegraphics[width=0.3cm]{figures/logo/iron_ingot.pdf}} & \multicolumn{2}{c}{Golden Shovel \includegraphics[width=0.3cm]{figures/logo/Golden_Shovel.pdf}} & \multicolumn{2}{c}{Diamond Sword \includegraphics[width=0.3cm]{figures/logo/diamond_sword.pdf}} & \multicolumn{2}{c}{Avg} \\
\cmidrule(lr){2-3} \cmidrule(lr){4-5} \cmidrule(lr){6-7} \cmidrule(lr){8-9} \cmidrule(lr){10-11} \cmidrule(lr){12-13}
 & \texttt{CR} & \texttt{SR}  & \texttt{CR} & \texttt{SR}  & \texttt{CR} & \texttt{SR}  & \texttt{CR} & \texttt{SR}  & \texttt{CR} & \texttt{SR}  & \texttt{CR} & \texttt{SR} \\
\hline
\multicolumn{13}{c}{\textit{Multimodal Large Language Model$^{\dagger}$}} \\
\hline
GPT-5-Instant \cite{gpt5}             & 0.22 & 0.05 & 0.18 & 0.05 & 0.14 & 0.00 & 0.12 & 0.00 & 0.12 & 0.00 & 0.15 & 0.02 \\
GPT-5-Thinking \cite{gpt5}             & 0.28 & 0.10 & 0.24 & 0.05 & 0.16 & 0.00 & 0.14 & 0.00 & 0.12 & 0.00 & 0.18 & 0.03 \\
Gemini-2.5-Flash \cite{gemini2.5}     & 0.24 & 0.10 & 0.22 & 0.10 & 0.16 & 0.05 & 0.14 & 0.05 & 0.14 & 0.00 & 0.18 & 0.06 \\
Gemini-2.5-Pro \cite{gemini2.5}     & 0.36 & 0.25 & 0.30 & 0.20 & 0.28 & 0.10 & 0.22 & 0.05 & 0.14 & 0.00 & 0.26 & 0.12 \\
Qwen2.5-VL-7B \cite{bai2025qwen2}   & 0.31 & 0.20 & 0.26 & 0.20 & 0.24 & 0.10 & 0.19 & 0.05 & 0.09 & 0.00 & 0.21 & 0.11 \\
\hline
\multicolumn{13}{c}{\textit{Agents in Minecraft}} \\
\hline
Voyager$^{*}$ \cite{wang2023voyager} \texttt{[NeurIPS'23]}   & 0.16 & 0.10 & 0.16 & 0.05 & 0.14 & 0.00 & 0.14 & 0.00 & 0.08 & 0.00 & 0.14 & 0.03 \\
DEPS$^{*}$ \cite{wang2023describe} \texttt{[NeurIPS'23]}   & 0.14 & 0.10 & 0.14 & 0.10 & 0.12 & 0.05 & 0.12 & 0.00 & 0.10 & 0.05 & 0.12 & 0.06 \\
MP5$^{*}$ \cite{qin2023mp5}  \texttt{[CVPR'24]}  & 0.29 & 0.20 & 0.19 & 0.20 & 0.18 & 0.15 & 0.18 & 0.05 & 0.16 & 0.10 & 0.20 & 0.14 \\
Jarvis-1$^{*}$ \cite{wang2023jarvis} \texttt{[TPAMI'25]}   & 0.19 & 0.15 & 0.18 & 0.15 & 0.18 & 0.10 & 0.20 & 0.10 & 0.12 & 0.05 & 0.17 & 0.11 \\
Optimus-1$^{*}$ \cite{li2024optimus} \texttt{[NeurIPS'24]}   & 0.21 & 0.20 & 0.18 & 0.20 & 0.18 & 0.15 & 0.17 & 0.15 & 0.12 & 0.10 & 0.17 & 0.16 \\
Optimus-2$^{*}$ \cite{li2025optimus}  \texttt{[CVPR'25]}   & 0.22 & 0.25 & 0.21 & 0.15 & 0.18 & 0.15 & 0.19 & 0.10 & 0.19 & 0.15 & 0.19 & 0.16 \\
\hline
\rowcolor[HTML]{E7EEFE}
Optimus-3           & 
\textbf{0.89} & \textbf{0.75} & \textbf{0.86} & \textbf{0.70} & \textbf{0.79} & \textbf{0.65} & \textbf{0.75} & \textbf{0.55} & \textbf{0.69} & \textbf{0.35} & \textbf{0.79} & \textbf{0.60} \\
\bottomrule[1.2pt]
\end{tabular}
}
\end{table*}
\noindent\textbf{Baselines}. For the MineSys2 Benchmark, we compare Optimus-3 against generalist MLLMs (GPT-4o \cite{gpt4}, Qwen2.5-VL \cite{bai2025qwen2}, Gemini-1.5-pro \cite{team2024gemini}), and various variants of Qwen2.5-VL \cite{bai2025qwen2} that post-trained on OptimusM$^4$ dataset.

\noindent\textbf{Results Analysis}.  As depicted in Table \ref{tb:main_text}, compared to all baselines, Optimus-3 achieves the highest performance across all System 2 tasks. The first four rows of Table \ref{tb:main_text} reveal the limited capabilities of existing generalist MLLMs, due to the fact that they have not been post-trained in the Minecraft domain. In contrast, \texttt{Qwen2.5-VL-SFT}, which post-trained on OptimusM$^4$, shows a notable performance boost. Furthermore, after applying our proposed DGRPO method, \texttt{Qwen2.5-VL-RL} achieves a 52\% improvement in \textit{Grounding}. More importantly, we observe that the various variants of Qwen2.5-VL exhibit task interference, while Optimus-3 consistently achieves superior performance across tasks. Moreover, compared to token-level routing, the proposed task-level routing demonstrates superior performance on tasks such as \textit{Captioning}, \textit{Planning}, and \textit{Grounding}. We attribute this improvement to the Dual-route Aligned MoE architecture in Optimus-3, which allocates task-specific experts via task router, thereby effectively mitigating conflicts among heterogeneous tasks. 

\begin{figure}[t]
\centering
\includegraphics[width=0.45\textwidth]{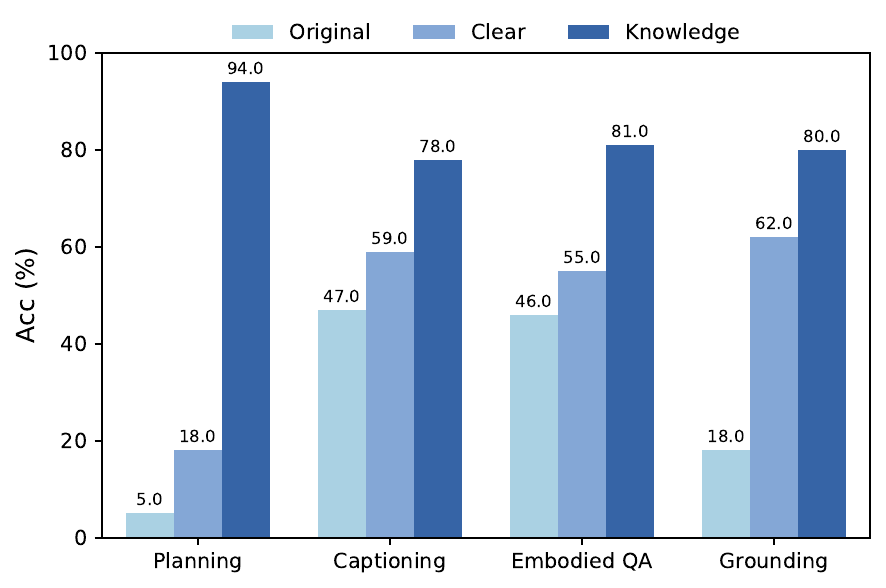}
\caption{Ablation Study on Training Data. \texttt{Original} refers to the original Qwen2.5-VL-7B, \texttt{Clear} indicates the Optimus-3 post-trained on data without knowledge, \texttt{Knowledge} represents the Optimus-3 post-trained on OptimusM$^4$.}
\label{fig:fig8}
\end{figure}

\subsection{Evaluation on System 1 Tasks}
\noindent\textbf{Evaluation Tasks \& Metrics}. To evaluate the robustness of Optimus-3 in executing long-horizon action sequences, we follow the experimental setup of prior work and conduct experiments on the Long-Horizon benchmark \cite{li2024optimus}. It comprises 67 tasks spanning 7 technology groups: Wooden (10 tasks), Stone (9 tasks), Iron (16 tasks), Gold (6 tasks), Diamond (7 tasks), Redstone (6 tasks), and Armor (13 tasks). Each task consists of multiple sub-steps that the agent must execute sequentially to reach the final goal. For each task, we perform at least 30 rollouts from different initial positions, and we report the average Success Rate (\texttt{SR}) with standard deviation on each task group as the evaluation metric.

\noindent\textbf{Baselines}. We follow the hierarchical architecture of agents from prior work, employing an MLLM (GPT-3.5 \cite{gpt3}, GPT-4o \cite{gpt4}, Qwen2.5-VL\cite{bai2025qwen2}, Gemini-1.5-pro \cite{team2024gemini}) as a planner to generate sub-goals, followed by STEVE-1 as a policy to sequentially generate actions. Moreover, we employ current SOTA agents in Minecraft (DEPS \cite{wang2023describe}, Jarvis-1 \cite{wang2023jarvis}, Optimus-1 \cite{li2024optimus}, and Optimus-2 \cite{li2025optimus}) as baselines.

\begin{figure}[t]
\centering
\includegraphics[width=0.4\textwidth]{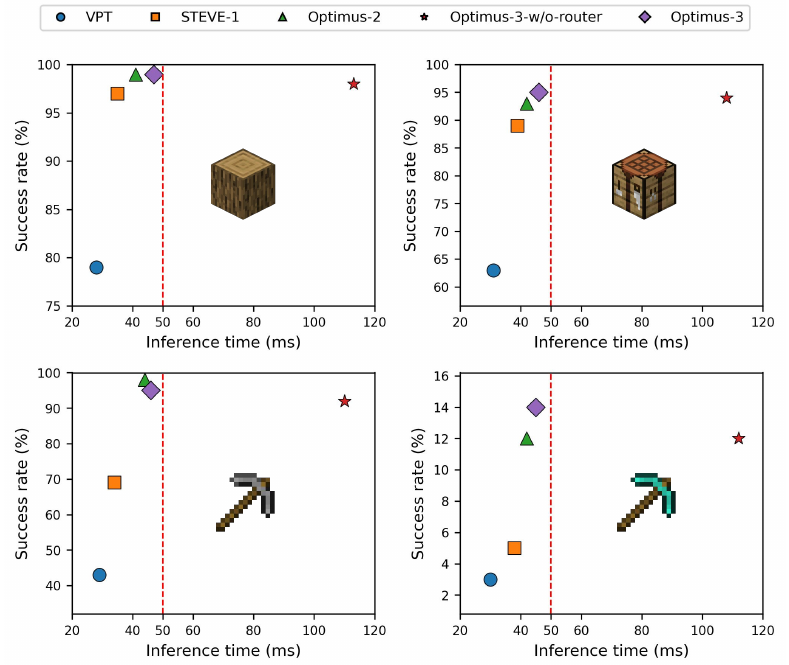}
\caption{Comparison of the average per-action inference time on Log \includegraphics[width=0.3cm]{figures/logo/wood.pdf}, Crafting Table \includegraphics[width=0.3cm]{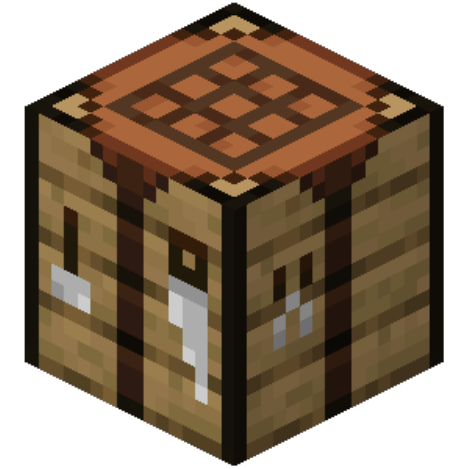}, Stone Pickaxe \includegraphics[width=0.3cm]{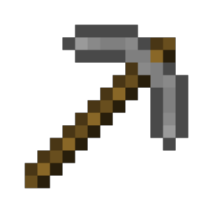}, and Diamond Pickaxe \includegraphics[width=0.3cm]{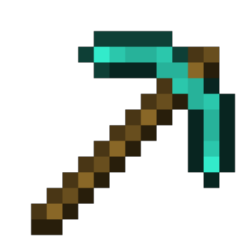}. \texttt{Optimus-3-w/o-router} denotes the variant of Optimus-3 without Layer Router. The 50\,ms mark corresponds to Minecraft's real-time interaction requirement, i.e., 20\,Hz.}
\label{fig:fig9}
\end{figure}
\noindent\textbf{Results Analysis}. As shown in Table \ref{tb:main_horizon}, Optimus-3 achieves the highest success rate across all 7 task groups, with particularly strong performance on the Diamond Group, attaining a \texttt{SR} of 15\%. It is worth emphasizing that Optimus-3 performs both high-level planning and low-level action execution within an end-to-end architecture, whereas all baseline agents \cite{wang2023describe,wang2023jarvis,li2024optimus,li2025optimus} rely on additional external tools or task-specific modules to bridge these two stages. The results in Rows 1-4 of Table~\ref{tb:main_horizon} show that generic MLLMs struggle to produce effective long-horizon plans, which we attribute to the absence of domain-specific fine-tuning in Minecraft. In contrast, \texttt{Qwen2.5-VL-SFT} is fine-tuned on our proposed OptimusM$^4$ dataset, its planning capability improves substantially. It highlights the critical role of OptimusM$^4$ in injecting Minecraft-specific knowledge and structured long-horizon planning skills. Moreover, compared with the variant (\texttt{Optimus-3-Action}) trained solely on action-trajectory data, Optimus-3 trained on a diverse set of task types achieves a substantially higher success rate. We attribute this improvement to the richer domain knowledge contained in the different task families, which enhances the model's representations. While shared knowledge expert in Optimus-3 enables such knowledge transfer across tasks.

\subsection{Evaluation on Open-ended Tasks}

\noindent\textbf{Evaluation Tasks \& Metrics}. To evaluate the performance of Optimus-3 to jointly deploy its diverse capabilities in open-ended scenarios, we construct a suite of five open-ended tasks corresponding to the Wooden, Stone, Iron, Gold, and Diamond tech levels. For each episode, the agent is randomly initialized at a different location, and endowed with a set of initial resources sampled from a predefined resource pool. This setup requires the agent to leverage its multi-dimensional abilities, e.g.,  perception, planning, and reflection, to make adaptive decisions conditioned on the current state. For each task, we perform 20 rollouts and employ the average Success Rate (\texttt{SR}) and average Completion Rate (\texttt{CR}) as evaluation metrics. \begin{figure*}[t]
\centering
\includegraphics[width=1\textwidth]{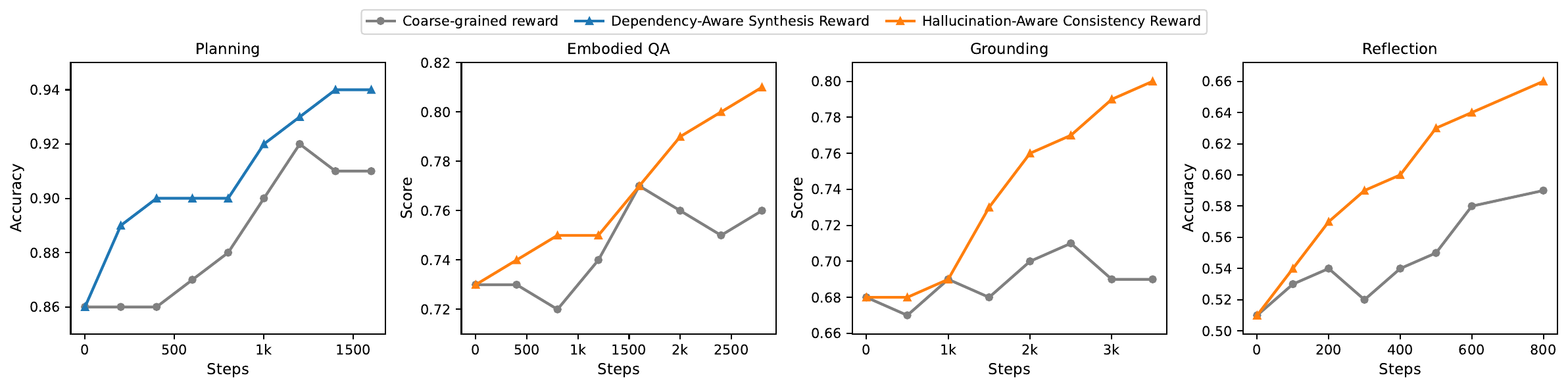}
\caption{Performance comparison between coarse-grained reward (GRPO) and our fine-grained rewards (DGRPO). Our Dependency-Aware Synthesis Reward and Hallucination-Aware Consistency Reward enable stable and consistent performance improvements, whereas using coarse-grained rewards leads to unstable RL training due to the lack of fine-grained supervision.
}
\label{fig:fig10}
\end{figure*}

\begin{table}[htbp]
\centering
\caption{
Ablation study of Optimus-3 on training stages. We report accuracy of Planning, Captioning, EQA, and Grounding on each stage.}
\label{tb:ablation_method}

\resizebox{0.48\textwidth}{!}{%
\begin{tabular}{ccc|cccc}
\toprule[1.2pt]
SFT & Coldstart & RL & Planning & Captioning & EQA & Grounding \\
\hline
          &           &           & 0.05 & 0.47 & 0.46 & 0.18 \\
\Checkmark &          &           & 0.79 & 0.68 & 0.71 & 0.52 \\
\Checkmark & \Checkmark &         & 0.86 & 0.72 & 0.73 & 0.68 \\
\Checkmark &          & \Checkmark & 0.76 & 0.71 & 0.68 & 0.79 \\

\rowcolor[HTML]{E7EEFE}
\Checkmark & \Checkmark & \Checkmark & \textbf{0.94} & \textbf{0.78} & \textbf{0.81} & \textbf{0.80} \\
\bottomrule[1.1pt]
\end{tabular}%
}
\end{table}
\texttt{CR} quantifies the agent's progress toward solving an open-ended task by measuring how many of the following stages it successfully completes in order: (1) Captioning, (2) Grounding, (3) Planning, (4) Action, and (5) Embodied QA. Formally, let $K$ denote the total number of stages and let $k_i$ be the number of stages completed in the $i$-th rollout. Given $N$ rollouts, the \texttt{CR} is defined as:
\begin{equation}
\texttt{CR} \;=\; \frac{1}{N}\sum_{i=1}^{N} \frac{k_i}{K}\;,
\end{equation}

\noindent\textbf{Baselines}. For the Open-ended Tasks, we compare Optimus-3 against generalist MLLMs (GPT-5 \cite{gpt4}, Qwen3-VL \cite{bai2025qwen2}, Gemini-2.5-pro \cite{team2024gemini}), and current SOTA agents in Minecraft (Jarvis-1 \cite{wang2023jarvis}, Optimus-1 \cite{li2024optimus}, and Optimus-2 \cite{li2025optimus}).

\noindent\textbf{Results Analysis}. As shown in Table~\ref{tb:main_open}, Optimus-3 achieves the highest task success rate and completion rate across all open-ended tasks. Notably, on the most challenging Diamond Sword task, Optimus-3 attains a success rate of 35\% and a completion rate of 69\%, demonstrating a substantial margin over all baselines. In contrast, generic MLLMs achieve consistently low success and completion rates on open-ended tasks, as they often fail to generate adaptive plans conditioned on the specific situation (e.g., the current inventory and surrounding environment). Meanwhile, even strong baselines such as Jarvis-1 \cite{wang2023jarvis} and Optimus-1 \cite{li2024optimus} struggle to generalize to open-ended settings. We attribute this to the fact that they are neither explicitly designed nor trained to support multi-dimensional abilities beyond planning and action execution. It highlights the advantage of Optimus-3: by integrating multi-dimensional capabilities into an end-to-end framework, it can better adapt to open-ended scenarios and solve tasks that require coherent perception--grounding--planning--execution--reflection loops.

\begin{figure*}[htbp]
    \centering
    \includegraphics[width=1\textwidth]{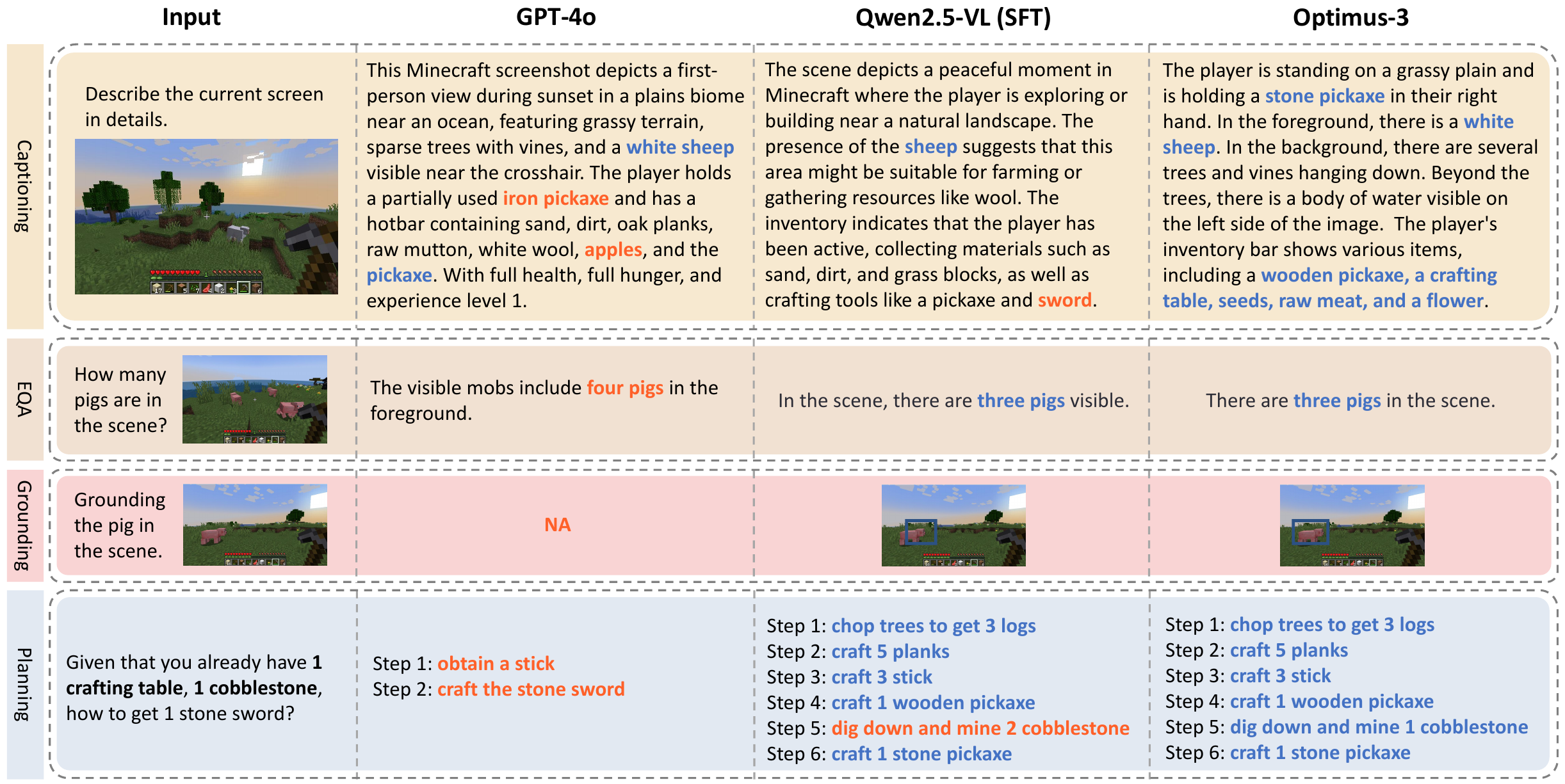}
    \caption{Visual comparison of Optimus-3 (ours), Qwen2.5-VL (tuned on our data), and GPT-4o. Red highlights indicate errors, while blue highlights denote correct outputs.}
    \label{fig:fig11}
\end{figure*}
\subsection{Ablation Study}
\label{ablation_study}
In this section, we conduct comprehensive ablation studies to validate the effectiveness of our approach and summarize our key findings.

\noindent\textbf{High-quality training data is essential for effective MLLM post-training}. Experimental results in Table \ref{tb:main_text} reveal that both Optimus-3 and \texttt{Qwen2.5-VL-SFT} benefit substantially from training on OptimusM$^4$ dataset. Furthermore, we conduct an ablation study to investigate the role of knowledge injection in our data-generation pipeline. As shown in Figure~\ref{fig:fig8}, removing the knowledge-augmentation mechanism (denoted as \texttt{Clear}) leads to substantial performance degradation, with drops of 81\% on Planning, 32\% on Embodied QA, and 23\% on Grounding, respectively. We attribute this degradation to the fact that generic MLLMs lack Minecraft-specific domain knowledge and thus produce low-quality synthetic data when the knowledge signal is removed. This result highlights the critical importance of our proposed knowledge-enhanced data generation pipeline in providing high-quality supervision for post-training the MLLM backbone. Moreover, the data collection process of OptimusM$^4$ incurred only \$300 in API costs, and was completed using 4$\times$ NVIDIA L40 GPUs over 36 hours, demonstrating the cost-efficiency of the pipeline.

\noindent\textbf{The Dual-Router Aligned MoE architecture is crucial for heterogeneous multi-task learning}. As shown in Table~\ref{tb:main_text}, dense models suffer from severe task interference when trained on heterogeneous multi-tasks, leading to suboptimal performance across all task types. Although the Optimus-3 variant with token routing outperforms the dense model on Planning and Embodied QA, it exhibits noticeable degradation on Captioning and Grounding. We attribute this behavior to the intrinsic difficulty of token-level routing, which must simultaneously address load balancing across experts and joint optimization over conflicting task objectives. In contrast, Optimus-3 with task routing achieves the best performance on all tasks. It indicates that explicitly assigning task-specific experts for heterogeneous tasks substantially mitigates task interference.

On the other hand, we conduct an ablation study to evaluate the impact of the Layer Router on latency-sensitive tasks, as illustrated in Figure~\ref{fig:fig9}. The results indicate that the Optimus-3 variant without the Layer Router suffers from an average per-action latency exceeding 50\,ms, thus failing to meet Minecraft's 20\,Hz interaction requirement. In contrast, introducing the Layer Router substantially accelerates the inference of Optimus-3 while preserving its high success rate. Remarkably, with the layer router enabled, the Optimus-3 with 6.8B parameters attains an inference speed comparable to Optimus-2 with 1.3B parameters. It highlights the role of the layer router in adapting the effective model depth to the task-dependent reasoning complexity. Taken together, these findings highlight the effectiveness of our proposed Dual-Router Aligned MoE architecture in simultaneously resolving conflicts among heterogeneous tasks and balancing inference complexity with real-time latency constraints.

\noindent\textbf{Dual-Granularity Reasoning-Aware Policy Optimization further enhances the agent’s capabilities}. As shown in Table \ref{tb:ablation_method}, removing the DGRPO method (both the \texttt{coldstart} and \texttt{RL} stages) leads to performance drops of 16\%, 13\%, 12\%, and 35\% on Planning, Captioning, EQA, and Grounding, respectively. These results highlight the pivotal role of DGRPO in enhancing the performance of Optimus-3 in dynamic and diverse scenarios. Furthermore, we conduct experiments to investigate the importance of our proposed fine-grained reward design. As shown in Figure~\ref{fig:fig10}, compared with using only the final answer correctness as a coarse feedback signal, our Dependency-Aware Synthesis Reward and Hallucination-Aware Consistency Reward yield consistently better performance on Planning, Embodied QA, Grounding, and Reflection. A critical ablation on the Embodied QA reveals that the model fails to converge when restricted to standard answer-level supervision. It shows the limitations of outcome-based RL in complex visual reasoning: the absence of process supervision renders the model vulnerable to visual hallucinations. In contrast, our DGRPO mitigates this issue by imposing strict penalties on non-existent entities generated within the reasoning chain. 


\begin{figure*}[htbp]
    \centering
    \includegraphics[width=1\textwidth]{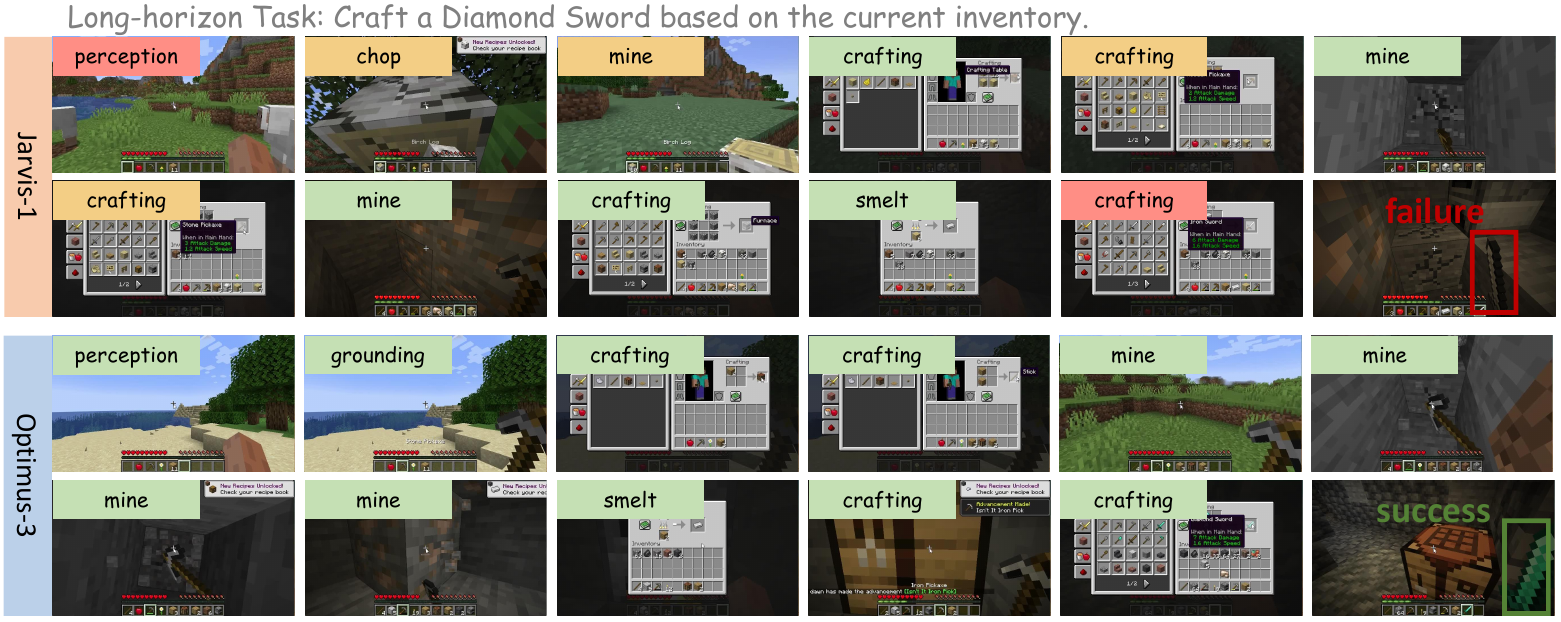}
    \caption{Visualization of Optimus-3 and Jarvis-1 executing the open-ended task \emph{``Craft a Diamond Sword based on the current inventory''}. Correct actions are highlighted in \colorbox{correct}{green}, unnecessary actions in \colorbox{unnec}{yellow}, and erroneous actions in \colorbox{wrong}{red}.}
    \label{fig:fig12}
\end{figure*}
\subsection{Qualitative Analysis} In this section, we present qualitative visualizations to illustrate the behavior of Optimus-3 across different task types. As depicted in Figure \ref{fig:fig11}, we provide a visual comparison between Optimus-3, Qwen2.5-VL-7B \cite{bai2025qwen2}, and GPT-4o \cite{gpt4}, highlighting their differences in performance and behavior across non-action tasks. We observe that GPT-4o exhibits hallucinations in captioning and embodied QA, lacks grounding capabilities, and produces unreasonable plans. In contrast, Qwen2.5-VL-SFT, which is fine-tuned on our OptimusM$^4$ dataset, shows reduced hallucination, acquires grounding and planning abilities, and generates more coherent outputs. Notably, Optimus-3 accurately performs vision-related tasks and produces well-structured plans conditioned on instructions, demonstrating its superior perception and reasoning in the Minecraft environment. 

Moreover, as shown in Figure~\ref{fig:fig12}, we compare Optimus-3 with Jarvis-1 \cite{wang2023jarvis} on the open-ended instruction, \textit{Craft a diamond sword based on the current inventory}. We observe that Jarvis-1 fails to accurately capture the fine-grained item information in the inventory (hotbar), which leads to planning trajectories containing unnecessary steps (e.g., crafting a Wooden Pickaxe \includegraphics[width=0.3cm]{figures/logo/wooden_pickaxe.pdf} and a Stone Pickaxe \includegraphics[width=0.3cm]{figures/logo/stone_pickaxe.pdf}). Moreover, the semantic gap between its planning and action modules causes it to eventually produce an incorrect target item. In contrast, Optimus-3 precisely understands the current context and generates an appropriate plan, successfully crafting the Diamond Sword \includegraphics[width=0.3cm]{figures/logo/diamond_sword.pdf} step-by-step. We attribute this performance to Optimus-3's integration of System 1 action loops and System 2 reasoning capabilities within a unified framework, which enables it adapts more effectively to the diverse situations encountered in Minecraft.

\section{Conclusion}

In this paper, we presented \textbf{Optimus-3}, a unified generalist agent that organically integrates System 1 action loops with System 2 reasoning capabilities within an end-to-end framework. To overcome the challenges of data scarcity, architectural conflict, and open-world generalization, we contributed advances along three dimensions. First, we introduced a \textbf{Knowledge-Enhanced Data Generation Pipeline} that samples high-fidelity System 2 reasoning traces from raw interaction trajectories. By leveraging domain constraints to filter hallucinations, we constructed and released the \textbf{OptimusM$^4$} dataset. 
Second, we proposed the \textbf{Dual-Router Aligned MoE} architecture to address the computational conflict between the two systems. Through horizontal parameter decoupling and vertical depth adaption, it efficiently maintains a ``Fast Path'' for reflexive control and a ``Deep Path'' for deliberative reasoning. 
Third, we developed the \textbf{Dual-Granularity Reasoning-Aware Policy Optimization (DGRPO)} algorithm. It establishes a Process-Outcome Co-Supervision mechanism, utilizing dual-granularity rewards to align reasoning chains with visual evidence. Extensive experiments demonstrate that Optimus-3 achieves superior performance across diverse tasks, marking a significant step toward achieving general-purpose embodied intelligence in complex, open-ended worlds.

\bibliographystyle{IEEEtran}
\bibliography{ref}

\end{document}